%% file: paper.tex
\documentclass[journal]{IEEEtran}
\IEEEoverridecommandlockouts

\usepackage{xcolor} 

\usepackage{algorithm}
\usepackage{algpseudocode}

\algnewcommand\algorithmicinput{\textbf{Input:}}
\algnewcommand\Input{\item[\algorithmicinput]}
\algnewcommand\algorithmicoutput{\textbf{Output:}}
\algnewcommand\Output{\item[\algorithmicoutput]}

\makeatletter
\let\OldStatex\Statex
\renewcommand{\Statex}[1][0]{%
  \setlength\@tempdima{\algorithmicindent}%
  \OldStatex\hskip\dimexpr#1\@tempdima\relax}
\makeatother

\definecolor{blue}{rgb}{0,0,1}

\usepackage{graphics} 
\usepackage{epsfig} 
\usepackage{amsmath} 
\usepackage{amssymb}  
\usepackage{amsmath,amssymb,amsthm,amsfonts}
\usepackage{graphicx}
\usepackage{textcomp}
\usepackage{xcolor}
\usepackage{xspace}
\usepackage{amsfonts}  
\usepackage{mathrsfs}  
\usepackage{float}
\usepackage{lettrine}
\usepackage{enumerate}

\usepackage[utf8]{inputenc}
\usepackage{booktabs}    
\usepackage{caption}     
\usepackage{graphicx} 
\usepackage{pifont}

\newtheorem{theorem}{Theorem}

\newtheorem{remark}{Remark}

\usepackage{dblfloatfix}  
\usepackage{placeins}     
\usepackage{wrapfig,lipsum,booktabs}
\usepackage{float}
\usepackage{placeins}  
\usepackage{balance}
\usepackage{afterpage}
\usepackage[
colorlinks,linkcolor=blue]{hyperref}
\hypersetup{
	colorlinks=true,       
	linkcolor=blue,        
	citecolor=blue,        
	filecolor=magenta,     
	urlcolor=blue         }
\usepackage{cite}
\usepackage{mathtools}
\usepackage{subcaption}
\def \R{\mathbb{R}}
\def \Sym{\mathbb{S}}
\def \PSD{\Sym_{+}}

\def \data{\tilde{z}}
\newcommand{\MLE}[1]{\hat{#1}_{\textnormal{MLE}}}
\def \opt{^\star}

\newcommand{\st}{\textnormal{s.t.\; }}

\input{define}

\def \transpose{^\mathsf{T}}
\def \inv{^{-1}}

\DeclareMathOperator{\rank}{rank}

\DeclareMathOperator{\Orthogonal}{O}
\DeclareMathOperator{\SO}{SO}
\DeclareMathOperator{\SE}{SE}

\DeclareMathOperator{\Stiefel}{St}

\def \Graph{\mathcal{G}}
\def \Variables{\mathcal{V}}
\def \Factors{\mathcal{F}}
\def \Edges{\mathcal{E}}
\def \ParamSpace{\mathcal{X}}
\def \LiftedParamSpace{\mathcal{Y}}

\newcommand{\lifted}[1]{\bar{#1}}

\DeclareMathOperator*{\argmin}{argmin} 

\newcommand{\factorind}{k} 
\newcommand{\numfactors}{K}

\newcommand{\constraintind}{m}
\newcommand{\numconstraints}{M}

\newcommand{\blockvarind}{n}
\newcommand{\numblockvar}{N}

\newcommand{\totalprobdim}{r}

\newcommand{\vardim}{d}
\newcommand{\rankrestriction}{p}

\theoremstyle{definition}
\newtheorem{problem}{Problem}

\begin{document}

\title{Certifiable Factor Graph Optimization}

\author{Zhexin Xu, Nikolas R.\ Sanderson, Hanna Jiamei Zhang, and David M.\ Rosen,~\IEEEmembership{Senior Member,~IEEE}%
\thanks
{The authors are with the Robust Autonomy Lab, Institute for Experiential Robotics, Northeastern University, 360 Huntington Ave, Boston, MA 02115, USA. \texttt{\{xu.zhex, sanderson.n, zhang.hanna, d.rosen\}@northeastern.edu}. We acknowledge the support of the Natural Sciences and Engineering Research Council of Canada (NSERC).  Nous remercions le Conseil de recherches en sciences naturelles et en génie du Canada (CRSNG) de son soutien.  This work was funded in part by MIT Lincoln Laboratory through Air Force Contract FA8702-15-D-0001.
}
}

\maketitle

\begin{abstract}

We show that the \emph{factor graph} and \emph{certifiable estimation} paradigms, which have thus far been treated as essentially independent in the literature, can be naturally synthesized into a unified framework for \emph{certifiable factor graph optimization} that combines the ease of use of the former with the strong performance guarantees of the latter.  \emph{Factor graphs} provide a convenient modular modeling language that enables practitioners to easily design and deploy a wide variety of high-performance robotic state estimation systems by composing simple reusable building blocks; however, inference in these models is typically performed using \emph{local} optimization methods that can converge to significantly suboptimal solutions, a serious concern in safety-critical applications.  Conversely, \emph{certifiable estimators} based upon convex relaxation are provably capable of recovering \emph{verifiably globally optimal} state estimates in many practical settings; however, the computational cost of solving the large-scale relaxations that they employ necessitates the use of specialized, structure-exploiting optimization methods that require substantial expertise and manual effort to implement, significantly hampering their practical deployment.  The key insight enabling our synthesis of these paradigms is that the core mathematical constructions used to develop certifiable estimators (Shor's relaxation and Burer-Monteiro factorization) inherit a factor graph structure from the original problem: applying these transformations to a QCQP-representable estimation task with an associated factor graph model yields a \emph{lifted} problem with \emph{identical} factor graph connectivity whose constituent variables and factors are simple one-to-one algebraic transformations (\emph{lifts}) of those appearing in the original QCQP's factor graph.  This correspondence enables the Riemannian Staircase methodology for certifiable estimation to be easily instantiated and deployed using the same mature, highly-performant factor graph libraries and workflows already ubiquitously employed throughout robotics and computer vision.  Experimental evaluation on a variety of pose graph optimization, landmark SLAM, and range-aided SLAM benchmarks demonstrates that our certifiable factor graph optimization methodology enables the implementation of certifiable estimators that are functionally equivalent to current state-of-the-art \emph{hand-designed}, \emph{problem-specific} methods, while dramatically reducing the required implementation effort from the order of \emph{months} to \emph{hours}.


\end{abstract}


\section{Introduction}
\IEEEPARstart{S}{\emph{tate}} \emph{estimation} is the problem of inferring the values of a set of unknown parameters of interest from (typically noisy) sensor data; this capability lies at the heart of mobile robotics, supporting such foundational functions as perception, planning, navigation, and control \cite{Thrun2008Probabilistic, Barfoot_2017}.   Canonical examples of state estimation problems in robotics include \emph{simultaneous localization and mapping} (SLAM)~\cite{cadena2016past,rosen2021advances} and \emph{structure from motion} (SfM)~\cite{schonberger2016structure}, among many others.  

State estimation problems arising in robotics and computer vision applications are typically formalized as \emph{maximum likelihood estimation} (MLE) or  \emph{maximum a posteriori} (MAP) estimation tasks under an assumed probabilistic generative model for the available sensor data \cite{Thrun2008Probabilistic,dellaert2017factor}.  This formulation is attractive for its generality and conceptual simplicity, as well as for the strong statistical performance guarantees that MLE and MAP estimation afford \cite{Cover2006Elements}.  

At present, \emph{factor graphs} provide the dominant paradigm for modeling and solving state estimation problems in robotics and computer vision \cite{dellaert2017factor}.  In brief, a factor graph is a probabilistic graphical model that encodes the decomposition of a joint probability distribution over a collection of random variables into a product of simpler factors, each of which depends only upon a small \emph{subset} of the variables; intuitively, the factor graph can be thought of as describing how to \emph{construct} the joint distribution by ``wiring together'' these factors. In practical estimation problems,  the factors are typically conditional distributions that model the generation of individual sensor observations (e.g.\ a single camera image, laser scan, or GPS reading).  Consequently, factor graphs provide a convenient, modular modeling language that enables practitioners to easily specify even complex, high-dimensional estimation tasks (such as SLAM or 3D visual reconstruction problems) by composing a small number of common variable types (e.g.\ points, rotations, and poses) and factors representing common measurement modalities (e.g.\ cameras, lasers, IMUs, etc.).  Moreover, given a factor graph model of a probabilistic inference task, it is straightforward to automatically synthesize and deploy an efficient local optimization method to solve the associated MLE or MAP estimation problem.  This enables practitioners to focus primarily on \emph{modeling} the task to be solved (i.e.\ \emph{what} to do), rather than on the details of designing efficient inference algorithms (i.e.\ \emph{how} to do it).  This modeling convenience and ease of use is a primary reason for the widespread adoption of factor graph-based modeling and inference libraries \cite{Dellaert2012GTSAM, kummerle2011g, agarwal2012ceres}, as it enables even non-expert practitioners to quickly construct models of complex robotic perception, state estimation, and sensor fusion tasks, and then automatically synthesize and deploy efficient MLE-based estimators directly from these models.    However, this convenience comes at the expense of \emph{reliability}: because inference in factor graphs is typically performed using fast \emph{local} (rather than \emph{global}) optimization, there is no guarantee that it will return the correct (i.e. \emph{globally} optimal) maximum likelihood estimate.  The practical consequence is that standard factor graph-based inference can be alarmingly brittle, often returning egregiously wrong state estimates \emph{without warning}, even when the underlying estimation problem is well-posed~\cite{rosen2021advances}.  This lack of reliability is a serious limitation of the current state of the art, especially in high-impact but safety-critical applications such as autonomous vehicles, drones, and aerospace systems.

Alternatively, \emph{certifiably correct} estimation algorithms have recently emerged as a powerful class of techniques for robustly solving challenging state estimation tasks \cite{Rosen2025SLAMTheory}.  These techniques are based upon \emph{convex relaxation} rather than local search: they proceed by constructing a convex (typically \emph{semidefinite}) \emph{approximation} of an MLE or MAP problem, and then (\emph{globally}) solving this convex surrogate to produce a point estimate.  Remarkably, a growing body of work has shown that this approach can efficiently recover \emph{exact}, \emph{globally optimal} solutions to challenging estimation tasks in many practical settings; moreover, when exactness occurs, it is possible to \emph{verify} this fact \emph{a posteriori} by constructing a \emph{certificate of optimality} for the recovered solution \cite{Rosen2025SLAMTheory}.  Certifiable estimation methods thus provide strong guarantees of correctness and optimality that traditional factor graph-based inference methods lack.  However, despite their tremendous potential, certifiable estimators remain difficult to design and deploy in practice. In particular, they often require solving large-scale semidefinite relaxations that are beyond the capabilities of general-purpose (e.g.\ interior-point) optimization algorithms \cite{Todd2001Semidefinite}, necessitating the use of more scalable \emph{custom-built}, \emph{structure-exploiting} techniques.  The design and implementation of these specialized semidefinite optimization methods currently requires substantial expertise in applied mathematics (particularly convex analysis and numerical optimization), along with significant problem-specific manual analysis, design, and implementation effort.  These challenges have thus far presented a serious obstruction to the more widespread practical adoption of certifiable estimation methods.

In this paper, we show that the factor graph and certifiable estimation paradigms described above, which have thus far been treated as essentially independent in the literature, can be naturally synthesized into a unified framework for \emph{certifiable factor graph optimization} that combines the ease of use of the former with the strong performance guarantees of the latter, thereby achieving the best of both worlds.  The key insight enabling our synthesis is that the core mathematical constructions underpinning certifiable estimators (\emph{Shor's relaxation} \cite{shor1987quadratic} and \emph{Burer-Monteiro factorization} \cite{burer2003nonlinear}) inherit a natural factor graph structure from the original problem.  Specifically, given an estimation problem formulated as a \emph{quadratically-constrained quadratic program} (QCQP) with an associated factor graph model, we show that its corresponding Burer-Monteiro-factored Shor relaxation inherits a factor graph representation with identical connectivity, in which the constituent variables and factors are simple one-to-one algebraic transformations (\emph{lifts}) of those appearing in the original factor graph.  This result implies that the Burer-Monteiro-factored Shor relaxations can be instantiated and \emph{locally} optimized using existing factor graph libraries augmented with {lifted} variants of standard variable and factor types.  By embedding these local lifted factor graph optimizations within the Riemannian Staircase \cite{boumal2016non}, our framework enables practitioners to easily design and deploy certifiably correct estimators using the same mature, highly-performant factor graph libraries and workflows already ubiquitously employed throughout robotics and computer vision, without requiring specialized expertise in convex optimization or custom solver development.  In so doing, our approach extends the factor graph abstraction itself to encompass certifiable estimation, enabling these estimators to be specified directly at the level of factor graph models.  Experimental evaluation on a variety of pose graph optimization, landmark SLAM, and range-aided SLAM benchmarks demonstrates that our approach produces certifiable estimators that are functionally equivalent to current state-of-the-art \emph{hand-designed}, \emph{problem-specific} methods, while dramatically reducing the required implementation effort from the order of \emph{months} to \emph{hours}.  More broadly, our proposed framework applies to a wide range of estimation tasks beyond the canonical examples considered here, including problems involving more general sensing modalities and formulations not previously addressed in the certifiable estimation literature; extensions in these directions are currently under development and will be reported separately in future work.

In summary, the primary contributions of this work are:
\begin{enumerate} 

    \item We prove that the Burer-Monteiro-factored Shor relaxation of a QCQP-representable estimation task with an associated factor graph model admits a factor graph representation with identical variable-factor adjacency, in which the constituent variables and factors are obtained via simple one-to-one algebraic transformations (\emph{lifts}) of those appearing in the original QCQP's factor graph.
        
    \item Leveraging the preceding result, we show how to design, implement, and deploy a wide variety of Riemannian Staircase-based certifiable estimators using existing factor graph software libraries augmented with \emph{lifted} variable and factor types.
        
    \item We provide explicit constructions of the lifts of several variable types (rotations, translations, and unit vectors) and factors (relative rotation and translation measurements and point-to-point ranges) commonly appearing in robotic navigation and spatial estimation tasks.

    \item We release an open-source C++ implementation of our certifiable factor graph optimization methodology integrated into GTSAM.\footnote{We will provide a link to our implementation following  peer review.}
\end{enumerate}

The remainder of this paper is organized as follows.  
Section \ref{notation_section} establishes notation, and Section \ref{related_work_section} surveys related work.  
Sections \ref{section:factor_graphs_section} and \ref{certifiable_estimation_review_section} review the factor graph and certifiable estimation paradigms (respectively) that form the foundation of our approach.  
Section \ref{Certifiable_factor_graph_optimization_section} presents our main technical contributions, including the development of our certifiable factor graph optimization framework.  
Section \ref{common_lifted_types_section} provides concrete constructions of several common lifted variable and factor types.  
Section \ref{Experimental_results_section} presents experimental evaluation on pose graph optimization, landmark SLAM, and range-aided SLAM benchmarks.  
Finally, Section \ref{conclusion_section} concludes with a summary of contributions and directions for future work.

\section{Notation}
\label{notation_section}

In this section we fix notation that will be useful throughout the remainder of the paper.  Given an integer $n > 0$, we write $[n] \triangleq \lbrace 1, \dotsc, n \rbrace$ to denote the set of integers from $1$ to $n$ (inclusive).  Similarly, given an $n$-tuple $x = (x_1, \dotsc, x_n)$ and a subset $S = \lbrace i_1, \dotsc, i_k \rbrace \subseteq [n]$ of size $k$ with $i_1 < \dotsb < i_k$, we define $x_S$ to be the $k$-tuple whose elements are indexed by the elements of $S$:
\begin{equation}
x_S \triangleq (x_{i_1}, \dotsc, x_{i_k}).
\end{equation}
We write $\R$ to denote the set of real numbers.  Similarly, given an integer $n > 0$ we denote by $\Sym^n$ and $\PSD^n$ the sets of symmetric and symmetric positive-semidefinite matrices of size $n \times n$ (respectively).  We write $\Orthogonal(d)$, $\SO(d)$, and $\SE(d)$ for the \emph{orthogonal}, \emph{special orthogonal}, and \emph{special Euclidean} groups acting on $d$-dimensional Euclidean space (respectively).  We will identify $\Orthogonal(d)$ and $\SO(d)$ with their realizations as the following matrix Lie groups:
\begin{align}
\Orthogonal(d) &\triangleq \lbrace R \in \R^{d \times d} \mid R\transpose R = I_d \rbrace, \label{orthogonal_group_definition} \\
\SO(d) &\triangleq \lbrace R \in \R^{d \times d} \mid R\transpose R = I_d, \: \det(R) = +1 \rbrace,
\end{align}
and $\SE(d)$ with its realization as the semidirect product $\SE(d) \cong \R^d \rtimes \SO(d)$, with group operations:
\begin{subequations}
\begin{equation}
(t_1, R_1) \cdot (t_2, R_2) = (t_1 + R_1 t_2, \: R_1 R_2),
\end{equation}
\begin{equation}
(t,R)\inv = (-R\inv t, R\inv).
\end{equation}
\end{subequations}
Similarly, given integers $k$ and $n$ with $1 \le k \le n$, we let
\begin{equation}
\label{Stiefel_manifold_definition}
\Stiefel(k,n) \triangleq \lbrace X \in \R^{n \times k} \mid X\transpose X = I_k \rbrace 
\end{equation}
denote the set of ordered orthonormal $k$-frames in $n$-dimensional Euclidean space; this set forms a smooth compact matrix manifold called the \emph{Stiefel manifold} \cite{boumal2023introduction}.  Finally, we write $S^n$ to denote the $n$-dimensional unit sphere centered on the origin in $(n+1)$-dimensional Euclidean space:
\begin{equation}
\label{unit_sphere_definition}
S^n \triangleq \lbrace x \in \R^{n+1} \mid \lVert x \rVert_2 = 1 \rbrace.
\end{equation}
Note that comparing \eqref{orthogonal_group_definition} and \eqref{unit_sphere_definition} with \eqref{Stiefel_manifold_definition} reveals that the orthogonal group and the unit sphere are themselves specific instances of Stiefel manifolds:
\begin{equation}\label{stiefel_manifold_definition}
S^n = \Stiefel(1, n+1), \quad \quad \Orthogonal(n) = \Stiefel(n,n).
\end{equation}

\section{Related work}
\label{related_work_section}

In this section, we review prior work on two major paradigms for solving state estimation problems in robotics and computer vision: \emph{factor graph optimization} \cite{dellaert2017factor} and \emph{certifiable estimation} \cite{Rosen2025SLAMTheory}. Our primary focus is on algorithmic frameworks and computational methods for implementing these approaches at scale. 


At present, factor graphs comprise the dominant paradigm for modeling and solving state estimation problems in robotics and computer vision \cite{dellaert2017factor}.  These models provide a simple, modular abstraction that enables practitioners to specify a wide range of state estimation tasks by composing a relatively small library of standard variable and measurement types.  Moreover, given a factor graph model of a state estimation task, it is straightforward to algorithmically synthesize and deploy a fast \emph{local} optimization algorithm to solve the associated maximum likelihood estimation problem \cite{nocedal2006numerical}.  This factor graph-based modeling and optimization framework is the core mathematical and algorithmic abstraction underpinning several widely adopted software libraries, including GTSAM \cite{Dellaert2012GTSAM}, iSAM \cite{kaess2008isam, kaess2012isam2}, g2o \cite{kummerle2011g}, and Ceres \cite{agarwal2012ceres}, that are commonly used to implement state-of-the-art real-time robotic perception and state estimation systems in practice.  However, because inference in factor graphs is typically performed using \emph{local} (rather than \emph{global}) optimization, it is susceptible to convergence to significantly suboptimal local minima; that is, there is no guarantee of recovering the \emph{correct} (i.e.\ \emph{globally} optimal) maximum likelihood estimate.  As a result, these methods can be surprisingly brittle in practice, often returning egregiously wrong estimates \emph{without warning}.

Alternatively, recent work has developed a new class of \textit{certifiably correct} estimation algorithms that are provably capable of recovering \emph{globally optimal} solutions to state estimation problems in robotics and computer vision in many practical settings~\cite{rosen2019se,Rosen2025SLAMTheory}. The main idea behind these methods is to replace the nonconvex {optimization} underpinning traditional factor graph inference with a \emph{convex} (typically \emph{semidefinite}) \emph{relaxation} of the maximum likelihood estimation problem, which can be directly solved to \emph{global} optimality.  A growing body of work has shown (both theoretically and empirically) that the solutions recovered from these relaxations are frequently \emph{exact}, \emph{globally optimal} solutions of the original MLE problem (provided that the measurement noise is not too large).  Moreover, whenever exactness occurs, it can be \emph{verified} \emph{a posteriori} by constructing a \emph{certificate of optimality} for the recovered estimate.

However, despite their strong performance guarantees, at present certifiable estimation methods are difficult to deploy in practice  due to the high computational cost of solving the large-scale convex (typically {semidefinite}) relaxations that they employ. Standard off-the-shelf (interior-point) semidefinite optimization methods \cite{mosek, Sturm1999SeDuMi,Toh1999SDPT3,Yamashita2003Implementation} do not scale well beyond problems with a few thousand dimensions \cite{Todd2001Semidefinite}; however, many foundational estimation problems arising in robotics and computer vision, such as pose graph optimization \cite{grisetti2011tutorial} and bundle adjustment \cite{agarwal2010bundle}, involve tens to hundreds of thousands of variables.  As a result, a direct application of standard SDP optimization methods is computationally intractable for many problems of interest.

To address this computational challenge, prior work has explored several strategies for improving the scalability of semidefinite optimization methods \cite{Majumdar2020Scalability}. One common approach is to replace the semidefinite program itself with simpler convex approximations, such as linear programs (LPs), second-order cone programs (SOCPs), or sum-of-squares hierarchies like DSOS and SDSOS~\cite{ahmadi2019dsos}. While this approach can significantly reduce computational complexity, it does not guarantee the recovery of an \emph{exact} solution to the original SDP, thereby sacrificing the global optimality and certifiability properties that make certifiable estimation so appealing.  Another widely employed strategy is \emph{chordal decomposition}, which exploits sparsity in the data matrices to decompose a large SDP into smaller interconnected subproblems, thereby reducing the size of the largest matrix decision variable (which is the primary factor determining the cost of SDP optimization) \cite{Vandenberghe2015Chordal}.  However, the resulting decomposed problem is typically solved using nonsmooth first-order optimization methods such as consensus ADMM \cite{Boyd2011ADMM} that have sublinear convergence rates, and can thus exhibit prohibitively slow convergence when applied to badly-conditioned problems.  Unfortunately, many fundamental estimation problems in robotics, including pose graph optimization and bundle adjustment, are known to be poorly conditioned, rendering such methods impractically slow in these settings.

\emph{Burer-Monteiro (BM) factorization} \cite{burer2003nonlinear, boumal2016non} has emerged as the most effective strategy for solving the semidefinite programs underpinning large-scale certifiable estimators.  The main idea of this approach is to exploit the existence of \emph{low-rank} solutions to an SDP by \emph{reparameterizing} its decision variable $Z \in \PSD^n$ using a rank-$p$ symmetric factorization of the form $Z = YY\transpose$ (where $Y \in \R^{n \times p}$), and then optimizing over $Y$ instead.  This reformulation reduces the original (convex, high-dimensional) semidefinite program to a \emph{nonconvex} but low-dimensional \emph{nonlinear program}, yielding substantial savings in both memory and computation.  Crucially, it also enables the use of fast and scalable second-order nonlinear programming techniques to search for the optimal low-rank factor $Y$ \cite{nocedal2006numerical}.  Finally, in order to guarantee the recovery of a \emph{globally} optimal low-rank factor $Y^*$, the resulting nonconvex problems are embedded within the \emph{Riemannian Staircase} meta-algorithm, which solves a \emph{sequence} of BM-factored problems of increasing rank $p$ until an optimal solution to the original SDP is recovered \cite{boumal2016non, rosen2020Scalable}.  This algorithmic strategy forms the basis of essentially all large-scale certifiable estimators for robotic state estimation that have appeared in the literature to date \cite{rosen2019se, briales2017cartan, holmes2023efficient, fan2020cpl, papalia2024certifiably, tian2021distributed,Han2025Building}. 

However, despite their strong performance, BM-factorization-based certifiable estimators remain difficult to design and deploy in practice.  While in principle one can apply general-purpose  nonlinear programming methods to the low-dimensional BM-factored problems, even mature and highly-optimized methods such as IPOPT \cite{Waechter2006Implementation} and KNITRO \cite{Byrd2006KNITRO} can struggle to scale to the BM-factored NLPs arising in high-dimensional robotic state estimation tasks \cite{rosen2020Scalable}. Consequently, most state-of-the-art certifiable estimators instead exploit the favorable geometric (smooth manifold) structure of the feasible sets arising in estimation tasks by employing (\emph{intrinsic}) \emph{smooth manifold} optimization methods; these techniques effectively behave like \emph{unconstrained} optimization methods, and thus scale much more effectively to high-dimensional settings \cite{absil2008optimization}.  Unfortunately, while this strategy is highly effective, it is also highly labor-intensive: implementing such a method requires practitioners to construct an initial SDP relaxation, derive the corresponding Burer-Monteiro factorization, analyze the \emph{global} smooth manifold geometry of the BM-factored problem's feasible set, design a Riemannian optimization algorithm that is custom-tailored to \emph{that specific} smooth manifold, and then embed this custom optimization algorithm within the Riemannian Staircase meta-algorithm.  Executing this process requires substantial specialized subject-matter expertise (in convex analysis, differential geometry, and numerical optimization) and significant manual analysis, design, and implementation effort, both of which constitute a serious obstacle to the more widespread adoption and deployment of certifiable estimators in practice.

In this paper, we show that factor graph optimization and certifiable estimation, which have thus far been developed as essentially independent paradigms,\footnote{We note that \emph{Shonan Rotation Averaging} \cite{dellaert2020shonan} also explored the use of factor graph optimization to implement a certifiable estimator. However, that work was focused on designing a certifiable estimator \emph{specifically} for the rotation averaging problem, rather than developing a general framework relating factor graph optimization and certifiable estimation. From a methodological standpoint, it employs a fundamentally different approach, exploiting the fact that Stiefel manifolds can be represented as quotient spaces of rotation groups in order to express the underlying optimization problem within a Lie group-based factor graph framework; this construction is tightly coupled to the geometry of Stiefel manifolds and rotation groups, and thus does not readily extend to more general classes of estimation tasks. In contrast, the synthesis we develop here provides a general, systematic framework for constructing certifiable estimators directly from factor graph models, and is applicable across a wide range of estimation tasks.} can be synthesized into a unified theoretical and algorithmic framework for \emph{certifiable factor graph optimization}.  The key insight enabling this synthesis is that the core constructions underpinning certifiable estimation, namely Shor relaxation and Burer-Monteiro factorization, naturally \emph{inherit} a factor graph structure from the original estimation task.  This result implies that BM-factorization-based certifiable estimators can be designed and deployed using the same factor graph-based modeling abstractions and software libraries already employed to implement current state-of-the-art robotic state estimation systems, thereby eliminating the need for detailed manual problem-specific analysis and custom-designed Riemannian optimization algorithms.  Our approach thus provides a general and systematic framework for constructing certifiable estimators directly from factor graph models, thereby combining the modeling flexibility and ease of use of factor graphs with the strong global optimality guarantees of certifiable estimation.

\section{Factor graph-based modeling and estimation}
\label{section:factor_graphs_section}

In this section we briefly review the use of \emph{factor graphs} for modeling and solving statistical estimation problems arising in robotic perception and state estimation tasks.  Readers are encouraged to consult the monograph \cite{dellaert2017factor} for additional detail.

\subsection{Modeling perception problems with factor graphs}


State estimation problems in robotics and computer vision applications are typically formalized as \emph{probabilistic inference} tasks \cite{Thrun2008Probabilistic}.   Concretely, let $X \triangleq (X_1, \dotsc, X_\numblockvar)$ denote a collection of $\numblockvar$ unknown parameters (i.e.\ the \emph{model}) whose values we would like to infer, and $\data \triangleq \lbrace \data_\factorind \rbrace_{\factorind = 1}^\numfactors$ be a set of $\numfactors$ available noisy sensor measurements.  We assume that each measurement $\data_\factorind$ is sampled (conditionally) independently from a probabilistic generative model according to:
\begin{equation}
\label{measurement_model}
\data_k \sim p_\factorind \left(\cdot | X_{S_\factorind} \right) \quad \quad \forall k \in [\numfactors],
\end{equation}
where parameter indices $S_\factorind \subseteq [\numblockvar]$ specify the \emph{subset} $X_{S_\factorind}$ of the model parameters $X$ on which the $k$-th observation $\data_k$ depends.  Intuitively, each generative model \eqref{measurement_model} describes the operation of an individual sensor (i.e., it describes how a given state $X_{S_\factorind}$ of the world gives rise to a distribution $\data_k \sim p_\factorind(\cdot | X_{S_\factorind})$ over possible noisy readings $\data_k$ that sensor might return).

Let us now make a few simple but important observations about the measurement model \eqref{measurement_model}.  First, note that \eqref{measurement_model} makes explicit the fact that in many real-world applications, {each individual measurement} $\data_k$ typically only depends upon a \emph{small subset} $X_{S_\factorind}$ of the total set of model parameters $X$ (i.e.\ $\lvert S_\factorind \rvert \ll \numblockvar$).\footnote{For example, while a visual mapping task may involve the joint estimation of thousands of camera poses and millions of 3D point positions, \emph{each individual observed image feature point} $\data_k$  only depends upon \emph{two} of these model parameters: the pose of the individual camera that captured \emph{that specific} point observation, and the 3D position of the imaged point.}   Second, the conditional independence of the measurements $\data_k$  implies that the joint likelihood $p(\data | X)$ of the complete model $X$ given all of the data $\data$  factors as the product of the individual measurement likelihoods in \eqref{measurement_model}:
\begin{equation}
\label{conditional_factorization_of_joint_likelihood_function}
p(\data | X) = \prod_{k = 1}^\numfactors p_\factorind(\data_k | X_{S_\factorind}).
\end{equation}

\begin{figure}[t] 
  \centering
  \includegraphics[width=0.9\linewidth]{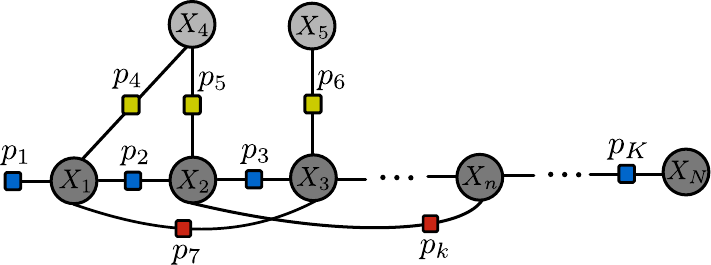} 
  \caption{An example factor graph.  This graph models a simple landmark SLAM problem containing $\numblockvar$ variables $X_\blockvarind$ and $\numfactors$ factors $p_\factorind$.  The variables consist of both \emph{robot poses} (dark gray) and \emph{landmarks} (light gray), while the factors consist of a unary \emph{prior} $p_1$ on the initial robot pose $X_1$, binary \emph{odometry} factors between successive robot poses (blue), \emph{landmark} observations (yellow), and \emph{loop closure} detections (red).}
  \label{fig:traj_overlay}
\end{figure}

\emph{Factor graphs} provide a convenient graphical representation for modeling the factorization \eqref{conditional_factorization_of_joint_likelihood_function} of the joint likelihood $p(\data | X)$.  Formally, the factor graph $\Graph$ (see Figure \ref{fig:traj_overlay}) associated to the factorization \eqref{conditional_factorization_of_joint_likelihood_function} is the bipartite graph $\Graph = (\Variables, \Factors, \Edges)$ in which:
\begin{itemize}
\item The \emph{variable nodes}  $\Variables \triangleq \lbrace X_1, \dotsc, X_\numblockvar \rbrace$ consist of the model parameters to be estimated;
\item The \emph{factor nodes} $\Factors \triangleq \lbrace p_1, \dotsc, p_\numfactors \rbrace$ consist of the individual factors (i.e. measurement likelihoods, conditional distributions) $p_\factorind$ appearing on the right-hand side of \eqref{conditional_factorization_of_joint_likelihood_function};
\item The \emph{edge set} is defined as:
\begin{equation}
\label{factor_graph_edge_set}
\Edges \triangleq \lbrace (X_\blockvarind, p_\factorind) \in \Variables \times \Factors \mid X_\blockvarind \in X_{S_\factorind} \rbrace;
\end{equation}
that is, variable $X_\blockvarind$ and factor $p_\factorind$ are joined by an edge in $\Graph$ if and only if $X_\blockvarind$ is an argument of the measurement likelihood $p_\factorind$ in \eqref{conditional_factorization_of_joint_likelihood_function}.
\end{itemize}

From a practical standpoint, factor graphs are useful because they provide a simple, general, modular modeling language for constructing complex, high-dimensional joint distributions $p(\data | X)$ by composing simple constituent parts [i.e.\ the individual measurement likelihoods $p_\factorind(\data_k | X_{S_\factorind})$ appearing in \eqref{measurement_model}].  Consequently, several current state-of-the-art software libraries for robotic perception \cite{gtsam,kummerle2011g,agarwal2012ceres} are based upon this abstraction: when using these libraries, a practitioner specifies a perception task by constructing its associated factor graph model.  This simple, object-oriented programming paradigm enables practitioners to easily model a vast array of robotic perception and state estimation tasks by combining standard building blocks, each corresponding to familiar state representations and sensing modalities.

\subsection{Maximum likelihood estimation in factor graphs}

Inference in factor graphs is typically performed via \emph{maximum likelihood estimation} (MLE), or more generally \emph{M-estimation} \cite{Huber2004Robust}.  Recall that in this approach, one determines a point estimate $\MLE{X}$ for the unknown parameter $X$ by \emph{maximizing} the value of the likelihood function $p(\data | X)$; intuitively, this is the point estimate that ``best explains" the available data $\data$.  Assuming that the likelihood function $p(\data | X)$ factors according to \eqref{conditional_factorization_of_joint_likelihood_function}, and that the $\blockvarind$-th parameter $X_\blockvarind$ takes values in the set $\ParamSpace_\blockvarind$, we may formalize the maximum likelihood estimation task as the following optimization problem:
\begin{equation} 
\label{factor_graph_MLE}
\MLE{X}(\data) \triangleq \argmin_{X_\blockvarind \in \ParamSpace_\blockvarind} \sum_{k = 1}^\numfactors \ell_\factorind(X_{S_\factorind}; \data_\factorind),
\end{equation}
where the $k$-th summand $\ell_\factorind(X_{S_\factorind}; \data_k)$ in \eqref{factor_graph_MLE} is either the negative log-likelihood $-\log p_\factorind(\data_k | X_{S_\factorind})$ of the $k$-th conditional distribution $p_\factorind$ in \eqref{conditional_factorization_of_joint_likelihood_function}, or a robust generalization thereof \cite{Huber2004Robust}.\footnote{While $\ell_{\factorind}$ models the \emph{negative log-likelihood} of the factor $p_\factorind$ in \eqref{factor_graph_MLE}, in the robotics literature it is common (in a slight abuse of terminology) to also refer to $\ell_{\factorind}$ itself as a ``factor", since this is ultimately the function that one actually implements in factor graph-based software libraries (c.f.\ e.g.\ \cite{Dellaert2012GTSAM,gtsam}) in order to instantiate the MLE \eqref{factor_graph_MLE}.  We will likewise adopt this convention.}

Formulating factor graph inference as the maximum likelihood estimation \eqref{factor_graph_MLE} provides several important advantages.  From a theoretical standpoint, maximum likelihood estimation provides strong guarantees on the statistical properties of the resulting estimator $\MLE{X}$, including asymptotic consistency and normality under mild conditions \cite{Ferguson1996Course}.  Computationally, in the typical case that the parameter spaces $\ParamSpace_\blockvarind$ are smooth manifolds and the summands $\ell_\factorind$ are smooth functions, the maximum likelihood estimation \eqref{factor_graph_MLE} is a smooth optimization problem defined over a product of smooth manifolds, enabling us to apply fast sparsity-exploiting first-or second-order smooth optimization methods on manifolds \cite{absil2008optimization,boumal2023introduction,nocedal2006numerical} to efficiently recover critical points.  This computational efficiency is essential for enabling real-time performance on resource-limited mobile robotic platforms (and indeed several state-of-the-art algorithms based upon this formulation are capable of processing inference problems involving tens to hundreds of thousands of states in real time \cite{rosen2014rise,kaess2012isam2,kummerle2011g}).  Finally, from an operational standpoint, in the typical case in which each parameter set $\ParamSpace_\blockvarind$ is a ``standard" smooth manifold whose global geometry and topology are known (e.g.\ common matrix Lie groups, etc.), it is possible to automatically (algorithmically) synthesize and run a fast smooth local optimization algorithm to recover a critical point of the MLE \eqref{factor_graph_MLE} directly from the factor graph model $\Graph$ describing the inference task.\footnote{More precisely:  we require that each $\ParamSpace_\blockvarind \subseteq \R^{\totalprobdim_\blockvarind}$ is a smooth embedded submanifold of some Euclidean space $\R^{\totalprobdim_\blockvarind}$, and that at each point $x \in \ParamSpace_\blockvarind$, we have access to computationally-convenient descriptions of (i) the tangent space $T_x(\ParamSpace_\blockvarind)$ of $\ParamSpace_\blockvarind$ at $x$, (ii) the orthogonal projection operator $\Pi_x \colon \R^{\totalprobdim_\blockvarind} \to T_x(\ParamSpace_\blockvarind)$ from the ambient space $\R^{\totalprobdim_\blockvarind}$ to the tangent space $T_x(\ParamSpace_\blockvarind) \subseteq \R^{\totalprobdim_\blockvarind}$ at $x$, and (iii) a \emph{retraction} operator $R_x \colon T_x(\ParamSpace_\blockvarind) \to \ParamSpace_\blockvarind$ at $x$ \cite{absil2008optimization,boumal2023introduction}.}  This capability is a core feature of several notable state-of-the-art software libraries for robotic perception (including GTSAM \cite{Dellaert2012GTSAM}, g2o \cite{kummerle2011g}, and Ceres \cite{agarwal2012ceres}), and is a major factor of their utility, as it enables practitioners to automatically synthesize and deploy highly-efficient inference algorithms directly from a factor-graph specification of the task to be solved, without requiring any specialized knowledge (e.g.\ of geometry or numerical optimization) or manual algorithm design on the part of the user.  The practical impact is that these libraries dramatically reduce the effort necessary to implement and deploy high-performance real-time perception systems, especially for users without extensive subject matter expertise in applied mathematics (e.g.\ differential geometry, numerical linear algebra, and numerical optimization).

However, despite its many favorable properties, practical implementations of the maximum likelihood estimation \eqref{factor_graph_MLE} often suffer from a lack of \emph{reliability}, due to their reliance upon fast \emph{local} optimization methods.  Specifically, these local optimization methods can only guarantee convergence to \emph{stationary points} of \eqref{factor_graph_MLE}, rather than the \emph{global minimizer} $\MLE{X}$ required in \eqref{factor_graph_MLE}.  The result is that the inference algorithms employed in common factor graph-based robotic perception libraries can be surprisingly brittle, often returning egregiously wrong estimates even when the problems to which they are applied are well-posed \cite{rosen2021advances}.  This lack of reliability motivates the use of the more robust \emph{certifiable} estimation methods we consider in the next section.

\section{Certifiable estimation}
\label{certifiable_estimation_review_section}

In this section, we briefly review the certifiable estimation pipeline that forms the foundation of our certifiable factor graph approach. Section~\ref{Shor_relaxation_section} introduces \emph{Shor's relaxation}, the convex relaxation upon which our certifiable estimators are built. Section~\ref{Burer_Monteiro_factorization_section} describes \emph{Burer-Monteiro factorization}, a computational technique for efficiently solving large-scale semidefinite programs by recasting them as lower-dimensional nonconvex programs, together with a verification procedure for certifying the global optimality of candidate solutions recovered from these lower-dimensional nonlinear programs. Finally, Section~\ref{Riemannian_Staircase_Section} presents the \emph{Riemannian Staircase}, a meta-algorithm that integrates these components into a complete procedure for efficiently recovering globally optimal solutions of large-scale SDP relaxations.

\subsection{Shor's relaxation}\label{Shor_relaxation_section}

Many common robotic perception and state estimation tasks have MLEs \eqref{factor_graph_MLE} that naturally take the form of (or can be equivalently recast as) \emph{quadratically-constrained quadratic programs} (QCQPs), i.e., optimization problems in which the objective and constraint functions are all \emph{quadratic}.  This class of problems admits a general procedure, called \emph{Shor's relaxation} \cite{shor1987quadratic}, for constructing strong convex relaxations.  

Let us assume that we have an estimation problem in the form of a standard QCQP: 
\begin{equation}
\label{QCQP_definition}
\begin{aligned}
f_{\text{QCQP}}\opt =  \min_{X \in \R^{\totalprobdim \times \vardim}} & \left\langle Q, X X\transpose\right\rangle \\
\text { s.t. } & \left\langle A_\constraintind, X X\transpose\right\rangle = b_\constraintind, \quad \constraintind\in[\numconstraints]
\end{aligned}
\end{equation}
with objective matrix $Q \in \Sym^\totalprobdim$, constraint matrices $A_\constraintind \in \Sym^\totalprobdim$ for all $\constraintind \in [\numconstraints]$, constraint vector $b \in \R^\numconstraints$, and $\vardim \le \totalprobdim$.  Note that the decision variable $X \in \R^{\totalprobdim \times \vardim}$ only enters problem \eqref{QCQP_definition} through the outer product $XX\transpose$, which is a symmetric, positive semidefinite matrix satisfying $\rank(XX\transpose) \le \vardim$ by construction.  

Shor's relaxation simply consists of replacing the low-rank symmetric outer product $XX\transpose$ appearing in the QCQP \eqref{QCQP_definition} with a \emph{generic} symmetric and positive semidefinite matrix $Z \in \PSD^\totalprobdim$ of the same dimension, producing the following (convex) \emph{semidefinite program} (SDP) relaxation:
\begin{equation}
\label{Shor_relaxation}
\begin{aligned}
f_{\text{SDP}}\opt = \min_{Z \in \PSD^\totalprobdim} & \langle Q, Z\rangle \\
\text { s.t. } & \left\langle A_\constraintind, Z\right\rangle=b_\constraintind, \quad \constraintind\in[\numconstraints].
\end{aligned}
\end{equation}

While simple to construct, Shor's relaxation \eqref{Shor_relaxation} turns out to possess many useful properties.  First, in marked contrast to the original nonconvex QCQP \eqref{QCQP_definition}, Shor's relaxation is a \emph{convex program}, from which we can recover a \emph{global} minimizer $Z\opt$ in polynomial time using standard interior-point methods \cite{potra2000interior}.  Second, it is easy to see that Shor's relaxation constitutes an \emph{outer approximation} of the original QCQP \eqref{QCQP_definition}: note that every point $X$ that is feasible for \eqref{QCQP_definition} maps to a symmetric outer product $XX\transpose$ that is also feasible for \eqref{Shor_relaxation}.  Intuitively, we can thus regard \eqref{Shor_relaxation} as being obtained from \eqref{QCQP_definition} by \emph{expanding the latter's feasible set}.  It immediately follows that the optimal value obtained by minimizing over this enlarged feasible set must \textit{lower bound} the optimal value of \eqref{QCQP_definition}:
\begin{equation}
\label{lower_bound_on_QCQP_optimal_value}
     f_{\text{SDP}}\opt \leq f_{\text{QCQP}}\opt.
\end{equation}

In turn, inequality \eqref{lower_bound_on_QCQP_optimal_value} enables us to easily \emph{upper bound} the suboptimality of \emph{any} feasible point $X$ as a solution of the QCQP \eqref{QCQP_definition} according to:
\begin{equation}
\label{suboptimality_upper_bound}
\left\langle Q, X X\transpose\right\rangle - f_{\text{QCQP}}\opt \le \left\langle Q, X X\transpose\right\rangle - f_{\text{SDP}}\opt.
\end{equation}
In particular, if the right-hand side of \eqref{suboptimality_upper_bound} is small, this enables us to conclude that $X$ is a \emph{near}-optimal solution of the QCQP \eqref{QCQP_definition} \emph{without} having to directly compute its optimal value $f_{\text{QCQP}}\opt$.  This observation forms the basis of the \emph{solution verification} mechanism used in certifiable estimators \cite{rosen2020Scalable}.  

Finally, suppose that we solve Shor's relaxation \eqref{Shor_relaxation}, and it so happens that the recovered minimizer $Z\opt \in \PSD^{\totalprobdim}$ satisfies $\rank(Z\opt) \le \vardim$.  Then $Z\opt$ admits a symmetric factorization of the form $Z\opt = {X\opt}{X\opt}\transpose$ for some $X\opt \in \R^{\totalprobdim \times \vardim}$, and it is straightforward to verify that the low-rank factor $X\opt$ so obtained is both feasible for the original QCQP \eqref{QCQP_definition} and achieves the same objective value as $Z\opt$.  Consequently, it follows [from \eqref{suboptimality_upper_bound}] that $X\opt$ is actually a \emph{globally} optimal solution for the original QCQP \eqref{QCQP_definition}.\footnote{In this case, the relaxation  \eqref{Shor_relaxation} is said to be \emph{exact}.}

The remarkable fact that justifies our interest in Shor's relaxation \eqref{Shor_relaxation} is that it is typically \emph{exact} when the original QCQP \eqref{QCQP_definition} is an instance of the maximum likelihood estimation \eqref{factor_graph_MLE}, provided that the noise on the available data $\data$ is not too large \cite{rosen2019se,cifuentes2022stability}.  In this favorable case, we can efficiently recover an \emph{exact, certifiably globally optimal} solution of the nonconvex QCQP (i.e., the \emph{correct} maximum likelihood estimate $\MLE{X}$) by solving the (\emph{convex}) Shor relaxation \eqref{Shor_relaxation}.  This algorithmic strategy is the basis of certifiable estimation methods.

\subsection{Solving the semidefinite relaxation}
\label{Burer_Monteiro_factorization_section}

In this section, we describe an effective algorithmic approach for solving large-scale instances of Shor's relaxation \eqref{Shor_relaxation} efficiently.

\subsubsection{Exploiting low-rank structure via Burer-Monteiro factorization}

As a semidefinite program, Shor's relaxation \eqref{Shor_relaxation} can \emph{in principle} be solved efficiently using general-purpose convex optimization algorithms (such as interior-point methods) \cite{boyd2004convex}; \emph{in practice}, however, the $\mathcal{O}(\totalprobdim^2)$ cost of storing and manipulating the dense symmetric matrix decision variable $Z$  prevents these general-purpose techniques from scaling to problems whose dimension $\totalprobdim$ is greater than a few thousand \cite{Todd2001Semidefinite}.  Unfortunately, many foundational robotic perception problems (such as SLAM) are often several orders of magnitude larger than this maximum effective problem size, placing them well beyond the reach of these general-purpose techniques.

On the other hand, the \emph{specific instances} of the SDP \eqref{Shor_relaxation} that we are interested in solving in the context of certifiable estimation exhibit additional useful structure: namely, they generically admit \emph{low rank solutions} $Z\opt$.   Indeed, we saw in the previous section that when Shor's relaxation \eqref{Shor_relaxation} of the QCQP \eqref{QCQP_definition} is exact, it admits a rank-$\vardim$ solution of the form $Z\opt = X\opt {X\opt}\transpose$, where $X\opt \in \R^{\totalprobdim \times \vardim}$ is a global minimizer of \eqref{QCQP_definition}.  More generally, a considerable body of prior work has shown that even instances of \eqref{Shor_relaxation} that are \emph{not} exact still admit solutions $Z\opt$ with a rank not much greater than $\vardim$.

In their seminal work, Burer and Monteiro \cite{burer2003nonlinear} proposed an elegant approach for efficiently solving large-scale SDPs that admit such low rank solutions.  Suppose that the SDP \eqref{Shor_relaxation} admits a solution $Z\opt \in \PSD^{\totalprobdim}$ of rank $\rankrestriction\opt \triangleq \rank(Z\opt)$.   Then $Z\opt$ can be concisely  represented using a symmetric rank factorization of the form $Z\opt = Y\opt {Y\opt}\transpose$ for some $Y\opt \in \R^{\totalprobdim \times \rankrestriction\opt}$.  Burer and Monteiro proposed to algorithmically exploit this observation by reparameterizing the decision variable $Z$ appearing in \eqref{Shor_relaxation} using an assumed rank-$\rankrestriction$ factorization of the form $Z = YY\transpose$.  Substituting this parameterization into Shor's relaxation then produces the following modification of \eqref{Shor_relaxation}, called the \emph{Burer-Monteiro} (BM) \emph{factorization}:
\begin{equation}
\label{rank_p_BM_factorization}
\begin{aligned}
f^\star_\text{BM} = \min _{Y \in \mathbb{R}^{\totalprobdim \times \rankrestriction}} & \left\langle Q, Y Y^{\top}\right\rangle \\
\text { s.t. } & \left\langle A_\constraintind, Y Y^{\top}\right\rangle=b_\constraintind, \quad \constraintind\in[\numconstraints].
\end{aligned}
\end{equation}

The BM factorization \eqref{rank_p_BM_factorization} affords several important advantages versus the original SDP \eqref{Shor_relaxation}.  First, the symmetric matrix decision variable $Z \in \PSD^{\totalprobdim}$ in \eqref{Shor_relaxation} requires $\mathcal{O}(\totalprobdim^2)$ storage, while the symmetric rank factor $Y \in \R^{\totalprobdim \times \rankrestriction}$ appearing in \eqref{rank_p_BM_factorization} requires only $\mathcal{O}(\totalprobdim \rankrestriction)$; in particular, this implies that the latter problem is \emph{much} lower-dimensional than the former when $\rankrestriction \ll \totalprobdim$.  Second, note that since $YY\transpose \succeq 0$ by construction, it is not necessary to explicitly enforce the positive-semidefiniteness constraint from the original SDP \eqref{Shor_relaxation} in its BM factorization \eqref{rank_p_BM_factorization}; consequently, the latter takes the form of a standard equality-constrained \emph{nonlinear program} (NLP), rather than a \emph{semidefinite program} that involves both equality and (more complex) conic positive-semidefiniteness constraints.  Finally, it is straightforward to verify that if the SDP \eqref{Shor_relaxation} admits a solution $Z\opt \in \PSD^{\totalprobdim}$ of rank $\rankrestriction\opt \triangleq \rank(Z\opt)$, then so long as the rank parameter $\rankrestriction$ in \eqref{rank_p_BM_factorization} is chosen such that $\rankrestriction \ge \rankrestriction\opt$, any (\emph{global}) minimizer $Y\opt \in \R^{\totalprobdim \times \rankrestriction}$ of \eqref{rank_p_BM_factorization} is a symmetric rank factor for a (global) minimizer ${Y\opt}{Y\opt}\transpose \in \PSD^{\totalprobdim}$ of \eqref{Shor_relaxation}.  Consequently, our approach will be to apply standard off-the-shelf (and highly mature, efficient, and scalable) nonlinear programming algorithms and software libraries to the BM factorization \eqref{rank_p_BM_factorization} in order to search for a low-rank factor $Y\opt$ of an optimal solution $Z\opt = {Y\opt}{Y\opt}\transpose$ of Shor's relaxation \eqref{Shor_relaxation}.

\subsubsection{Verifying global optimality}  While the BM factorization is amenable to fast \emph{local} optimization via standard NLP methods, the quadratic equality constraints appearing in \eqref{rank_p_BM_factorization} make this problem \emph{nonconvex}; consequently, there is no guarantee that a stationary point $Y\opt \in \R^{\totalprobdim \times \rankrestriction}$ recovered from \eqref{rank_p_BM_factorization} via fast \emph{local} optimization is a \emph{global} minimizer.  To address this potential pitfall, in this section we describe how to determine whether a given first-order stationary point $Y\opt \in \R^{\totalprobdim \times \rankrestriction}$ of the BM factorization \eqref{rank_p_BM_factorization} is a low-rank factor for a (\emph{global}) minimizer $Z\opt = {Y\opt}{Y\opt}\transpose$ of Shor's relaxation \eqref{Shor_relaxation}.

In brief, our optimality verification procedure is based upon comparing the first-order Karush-Kuhn-Tucker (KKT) conditions for the BM factorization \eqref{rank_p_BM_factorization} with those of the original SDP \eqref{Shor_relaxation}. To state these more concisely, let $\mathcal{A}: \Sym^\totalprobdim \to \R^\numconstraints$ denote the linear map defined by:
\begin{equation}
\mathcal{A}(Z)_\constraintind \triangleq \langle A_\constraintind, Z\rangle \quad \quad \forall m \in [\numconstraints],
\end{equation}
$\mathcal{A}^* \colon \R^{\numconstraints} \to \Sym^{\totalprobdim}$ its associated adjoint:
\begin{equation}
\label{adjoint_operator_definition}
\mathcal{A}^*(x) = \sum_{\constraintind = 1}^{\numconstraints} x_\constraintind A_\constraintind.
\end{equation}
With these definitions in hand, we may state the KKT conditions for \eqref{Shor_relaxation} and \eqref{rank_p_BM_factorization} as follows: 

\textbf{KKT conditions for Shor's relaxation:} A point $Z \in \PSD^{\totalprobdim}$ is a KKT point of \eqref{Shor_relaxation} if and only if there exists a corresponding Lagrange multiplier $\lambda \in \R^\numconstraints$ such that the following conditions are satisfied:
\begin{equation}
\label{KKT_conditions_for_SDP}
\mathcal{A}(Z) = b, \quad \quad SZ = 0, \quad \quad S \succeq 0,
\end{equation}
where $S\in \Sym^\totalprobdim$ is the symmetric matrix defined by:
\begin{equation}
\label{certificate_matrix_definition}
S \triangleq Q + \mathcal{A}^*(\lambda).
\end{equation}

\textbf{KKT conditions for the Burer-Monteiro factorization:}  A point $Y \in \R^{\totalprobdim \times \rankrestriction}$ is a KKT point of the BM factorization \eqref{rank_p_BM_factorization} if and only if there exists an associated Lagrange multiplier $\lambda \in \R^{\numconstraints}$ such that the following conditions are satisfied:
\begin{equation}
\label{KKT_conditions_for_BM_factorization}
\mathcal{A}(YY\transpose) = b, \quad \quad SY = 0,
\end{equation}
where $S$ is once again as defined in \eqref{certificate_matrix_definition}.

Comparing \eqref{KKT_conditions_for_SDP} and \eqref{KKT_conditions_for_BM_factorization},  we may deduce the following relation between KKT points of \eqref{Shor_relaxation} and \eqref{rank_p_BM_factorization}. 

\begin{theorem}[Theorem 4 of \cite{rosen2020Scalable}]
\label{verification_and_saddle_escape_theorem}
Suppose that Shor's relaxation \eqref{Shor_relaxation} satisfies Slater's constraint qualification, and let $Y \in \R^{\totalprobdim \times \rankrestriction}$ be a KKT point of the BM factorization \eqref{rank_p_BM_factorization} (with corresponding Lagrange multiplier $\lambda \in \R^{\numconstraints}$) that satisfies the linear independence constraint qualification.  Then exactly one of the following two cases holds:
\begin{enumerate}
\item [(a)]  $S \succeq 0$ and $Z = YY\transpose$ is a global minimizer of \eqref{Shor_relaxation}.
\item [(b)]There exists $v \in \R^{\totalprobdim}$ such that $v\transpose S v < 0$, and in that case, $Y_{+} \triangleq \left[Y \mid 0\right] \in \R^{\totalprobdim \times (\rankrestriction + 1)}$ is a KKT point of \eqref{rank_p_BM_factorization} attaining the same objective value as $Y$, and $\dot{Y}_{+} \triangleq \left[0 \mid v\right] \in \R^{\totalprobdim \times (\rankrestriction + 1)}$ is a feasible second-order direction of descent from $Y_{+}$.
\end{enumerate}
\end{theorem}

The practical significance of Theorem \ref{verification_and_saddle_escape_theorem} is that it provides an algorithmically useful \emph{theorem of the alternative}: under its hypotheses, a given stationary point $Y$ for the BM factorization \eqref{rank_p_BM_factorization} is either a low-rank factor for a solution $Z = YY\transpose$ of the original SDP \eqref{Shor_relaxation}, or we can construct a \emph{second-order direction of descent} $\dot{Y}_{+}$ from the embedding $Y_{+}$ of $Y$ into a higher-dimensional instance of \eqref{rank_p_BM_factorization}, enabling us to restart local optimization to continue the search for a low-rank factor of a solution of \eqref{Shor_relaxation}.

\subsection{Riemannian Staircase} \label{Riemannian_Staircase_Section}

Theorem \ref{verification_and_saddle_escape_theorem} immediately suggests a procedure  for efficiently recovering low-rank solutions of the SDP \eqref{Shor_relaxation} by applying standard local nonlinear optimization algorithms to a \emph{sequence} of progressively higher-dimensional instances of the Burer-Monteiro factorization \eqref{rank_p_BM_factorization}; this algorithm, called the \emph{Riemannian Staircase} \cite{boumal2015riemannian,boumal2016non}, is summarized as Algorithm \ref{Riemannian_Staircase_algorithm}.  In brief, starting at any initial feasible point $Y \in \R^{\totalprobdim \times \rankrestriction}$ for the $\rankrestriction$-dimensional instance of \eqref{rank_p_BM_factorization}, we apply \emph{local} optimization to recover a KKT point $Y^\star \in \R^{\totalprobdim \times \rankrestriction}$ and its associated Lagrange multiplier $\lambda^\star \in \R^{\numconstraints}$.  Given this stationary point, we then simply check [using the test prescribed in Theorem \ref{verification_and_saddle_escape_theorem}(a)] whether $Y^\star$ is a low-rank factor for a solution $Z^\star = {Y^\star}{Y^\star}\transpose$ of the SDP \eqref{Shor_relaxation}: if so, we return $Y^\star$, together with the optimal value $f_{\mathrm{SDP}}^{*}$ of the SDP; otherwise, we increase the dimension $\rankrestriction$ of the BM factorization \eqref{rank_p_BM_factorization}, and then invoke Theorem \ref{verification_and_saddle_escape_theorem}(b) to construct a \emph{second}-order direction of descent from the current estimate, enabling us to continue the search (via local optimization) for an optimal low-rank factor.

\begin{algorithm}[t]
\caption{The Riemannian Staircase}\label{Riemannian_Staircase_algorithm}
\begin{algorithmic}[1]
\Input Initial feasible point $Y \in \mathbb{R}^{\totalprobdim \times p}$ for $\rankrestriction$-dimensional instance of the Burer-Monteiro factorization (\ref{rank_p_BM_factorization}). 
\Output Symmetric factor $Y\opt$ for a minimizer $Z\opt = Y\opt {Y\opt}\transpose$ of SDP \eqref{Shor_relaxation}, optimal value $ f_{\mathrm{SDP}}^{*}$.
\Function{RiemannianStaircase}{$Y$}
\Loop 
\Statex[2]{\textcolor{blue}{// Find first-order KKT point of \eqref{rank_p_BM_factorization}.}}
\State $(Y^\star, \lambda^\star)$ $\gets$ \Call{LocalOptimization}{$Y$}.  \label{BM_local_opt_step}
\State Construct certificate matrix $S$ from $\lambda^\star$ as in \eqref{certificate_matrix_definition}.
\State $(\lambda_{\min}, v_{\min})$ $\gets$ \Call{MinimumEigenpair}{$S$}.
\Statex[2]{\textcolor{blue}{// Test optimality of $Z^\star = Y^\star {Y^\star}\transpose$ for \eqref{Shor_relaxation}.}}
\If{$\lambda_{\min} \ge 0$} 
\Statex[3]{\textcolor{blue}{// $Z^\star = Y^\star {Y^\star}\transpose$ minimizes \eqref{Shor_relaxation} by Thm.~\ref{verification_and_saddle_escape_theorem}(a).}}
\State $f_{\mathrm{SDP}}^{*} \gets \langle Q, Y^\star {Y^\star}\transpose\rangle$.
\State \Return $\lbrace Y^\star,  \: f_{\mathrm{SDP}}^{*}\rbrace$
\EndIf
\Statex[2]{\textcolor{blue}{// Construct lifted embedding of $Y^\star$.}}
\State $Y_{+} \gets \left[Y^\star \mid 0\right]$
\Statex[2]{\textcolor{blue}{// Construct second-order descent direction.}}
\State $\dot{Y}_{+} \gets \left[0 \mid v_{\min}\right]$
\Statex[2]{\textcolor{blue}{// Backtracking line search for saddle escape.}}
\State $Y$ $\gets$ \Call{LineSearch}{$Y_{+}, \dot{Y}_{+}$}
\EndLoop
\EndFunction
\end{algorithmic}
\end{algorithm}

\begin{remark}[Feasibility in local optimization in the Riemannian Staircase]
 For simplicity of presentation, the formulation of the Riemannian Staircase displayed in Algorithm \ref{Riemannian_Staircase_algorithm} implicitly assumes that the local optimization in line~\ref{BM_local_opt_step} always returns a feasible point for the BM-factored SDP \eqref{rank_p_BM_factorization}.  This assumption is justified whenever the feasible set for \eqref{rank_p_BM_factorization} turns out to be a smooth manifold, since this admits the use of specialized (smooth manifold-based) optimization algorithms that directly enforce feasibility throughout by construction \cite{boumal2023introduction,absil2008optimization}.  This is the case for the majority of previous applications of the Riemannian Staircase methodology in the certifiable estimation literature (cf.\ e.g.\ \cite{papalia2024overview} and the references therein), and is consistent with our usage of it in the example applications that we consider in Section \ref{Experimental_results_section}.  We note, however, that more general formulations of the Riemannian Staircase exist that admit the use of standard (extrinsic) constrained nonlinear programming techniques, which are \emph{not} guaranteed to recover a feasible point in all cases \cite{rosen2020Scalable}. All of the results presented in this paper extend immediately to this more general setting, provided that the corresponding generalization of Algorithm \ref{Riemannian_Staircase_algorithm} (cf.\ Algorithm 1 of \cite{rosen2020Scalable}) is employed.
\end{remark}

\section{Certifiable factor graph optimization}
\label{Certifiable_factor_graph_optimization_section}

In this section, we show how the factor graph and certifiable estimation paradigms described in Sections~\ref{section:factor_graphs_section} and~\ref{certifiable_estimation_review_section} can be naturally {synthesized} into a unified framework for \emph{certifiable factor graph optimization}. The key discovery enabling this synthesis is that factor graph structure is \emph{preserved} under Shor's relaxation and Burer-Monteiro factorization: given an initial QCQP with a nontrivial factor graph representation, we show that the Burer-Monteiro factorization of the associated Shor relaxation admits a factor graph representation with \emph{identical} connectivity, and in which the variables and factors are simple one-to-one algebraic transformations (\emph{lifts}) of the variables and factors appearing in the original QCQP's factor graph.  This result implies that the local optimizations required in each level of the Riemannian Staircase can be instantiated and run using current state-of-the-art factor graph optimization libraries, augmented with these lifted (i.e.\ \emph{certifiable}) variable and factor types.  The practical consequence is that our framework enables practitioners to easily design and deploy certifiable estimators using existing state-of-the-art factor graph-based libraries and workflows, dramatically lowering the barrier to entry and democratizing access to this powerful class of estimation methods.

\begin{figure*}[t]
  \centering
  \includegraphics[width=\textwidth]{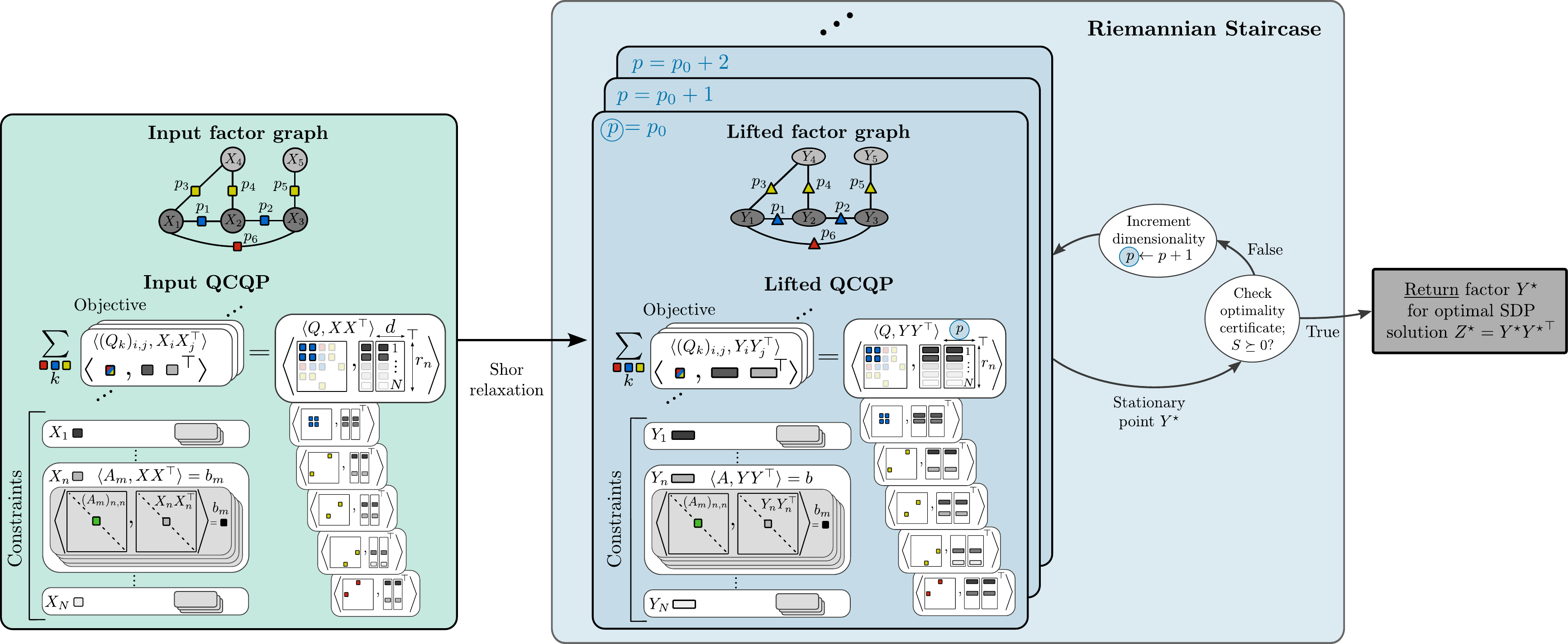}

\caption{Overview of our framework for certifiable factor graph optimization.  Starting from a factor graph model of a QCQP-representable estimation problem (left), we construct Shor's semidefinite relaxation and solve it using the {Riemannian Staircase} meta-algorithm (right).  In each iteration of the Staircase, we apply \emph{local} optimization to a Burer-Monteiro-factored instance of the relaxation to recover a first-order stationary point $Y^\star$.  We then test whether $Y^\star$ determines an optimal SDP solution $Z^\star = Y^\star {Y^\star}\transpose$ using the certificate matrix $S$: if $S \succeq 0$, then $Z^\star$ is certified optimal and the algorithm returns the symmetric factor $Y^\star$; otherwise, the factorization dimension $p$ is increased, and optimization resumes at the next level.  Crucially, Shor's relaxation and all of its associated Burer-Monteiro factorizations are parameterized by the same data matrices $\lbrace Q_k \rbrace$ and $\lbrace A_m \rbrace$ that define the original QCQP; in particular, this implies that each of the BM-factored SDP relaxations inherits a factor graph representation whose variables and factors are simple one-to-one algebraic transformations (\emph{lifts}) of those appearing in the input graph.  This enables us to \emph{automatically instantiate and locally optimize} the BM-factored SDP relaxations appearing in each level of the Riemannian Staircase directly from the input factor graph using existing factor graph software libraries.  Our approach thus preserves the modeling convenience and computational efficiency of current state-of-the-art factor graph-based software libraries and workflows while additionally providing the robustness and global optimality guarantees of certifiable estimation.}

  \label{Fig: framework}
\end{figure*}

\subsection{Factor graph structure in QCQPs}
\label{factor_graph_structure_in_QCQPs_subsection}

We begin the development of our certifiable factor graph framework by studying the natural setting for this synthesis: estimation problems that simultaneously admit a QCQP representation of the form \eqref{QCQP_definition} and a nontrivial factor graph representation of the form \eqref{factor_graph_MLE}.  Our goal is to elucidate how these two structures relate to each other. Our main result is that there is a precise correspondence between the structure of the factor graph and specific block sparsity and separability patterns  in the data matrices defining the QCQP \eqref{QCQP_definition}.


\subsubsection{Factor graph structure implies block sparsity and separability in QCQP data matrices}
\label{factor_graph_structure_implies_block_sparsity_and_separability_subsubsection}
Consider a maximum likelihood estimation problem in the form of a QCQP \eqref{QCQP_definition}, and suppose that the decision variable $X \in \mathbb{R}^{\totalprobdim\times d}$ is partitioned into $\numblockvar$ block rows:
\begin{equation}
\label{variable_block_decomposition}
    X \;=\;
    \begin{bmatrix} 
    X_1 \\ 
    \vdots \\ 
    X_\numblockvar 
    \end{bmatrix}
\end{equation}
(with $X_{\blockvarind} \in \R^{\totalprobdim_\blockvarind \times d}$ for all $\blockvarind \in [\numblockvar]$ and $\totalprobdim = \sum_{\blockvarind = 1}^\numblockvar \totalprobdim_\blockvarind$) such that the resulting MLE admits a factor graph model of the form $\Graph = (\Variables, \Factors, \Edges)$, where:
\begin{subequations}
\label{explicit_factor_graph_model_for_QCQP}
\begin{align}
\Variables &\triangleq \lbrace X_1, \dotsc, X_\numblockvar \rbrace, \label{variable_set_for_QCQP_factor_graph} \\
\Factors &\triangleq \lbrace \ell_1, \dotsc, \ell_\numfactors \rbrace, \label{factor_set_for_QCQP_factor_graph} \\
\Edges &\triangleq \lbrace (X_\blockvarind, \ell_\factorind) \in \Variables \times \Factors \mid \blockvarind \in S_\factorind \rbrace. \label{edge_set_for_QCQP_factor_graph}
\end{align}
\end{subequations}
Note that here the factors $\ell_\factorind$ are the (generalized) negative log-likelihoods defined in \eqref{factor_graph_MLE}, and $S_k \subseteq [\numblockvar]$ is the set of indices of the variables $X_1, \dotsc, X_\numblockvar$ that factor $\ell_\factorind$ depends upon for all $\factorind \in [\numfactors]$.  We will show that the structure \eqref{explicit_factor_graph_model_for_QCQP} of the factor graph $\Graph$ imposes specific block sparsity and separability patterns on the data matrices determining the QCQP \eqref{QCQP_definition}.

To see this, let us first consider the objective of the MLE [cf.\ \eqref{factor_graph_MLE}].  Under the stated hypotheses, each factor $\ell_\factorind$ in \eqref{factor_set_for_QCQP_factor_graph} is a quadratic function of the form:
\begin{equation}
\label{parameterization_of_quadratic_loss_function_in_factor_graph_objective}
\ell_\factorind(X) \triangleq \langle Q_\factorind, XX\transpose \rangle
\end{equation}
for some $Q_\factorind \in \Sym^{\totalprobdim}$.   Partitioning $Q_\factorind$ conformally  with \eqref{variable_block_decomposition}:
\begin{equation}
\label{variable_block_decomposition_of_objective_matrix}
Q_\factorind = 
\begin{bmatrix}
(Q_\factorind)_{1,1} & \cdots & (Q_{\factorind})_{1,\numblockvar} \\
 \vdots &  \ddots & \vdots \\
 (Q_\factorind)_{\numblockvar, 1} & \cdots & (Q_\factorind)_{\numblockvar, \numblockvar}
\end{bmatrix}
\end{equation}    
(where $(Q_\factorind)_{i,j} \in \R^{\totalprobdim_i \times \totalprobdim_j}$ for all $i,j \in [\numblockvar]$), we may substitute the block decompositions \eqref{variable_block_decomposition} and \eqref{variable_block_decomposition_of_objective_matrix} into \eqref{parameterization_of_quadratic_loss_function_in_factor_graph_objective}  to derive:
\begin{equation}
\label{block_decomposition_of_quadratic_factor}
\ell_\factorind(X) = \sum_{i,j = 1}^{\numblockvar} \left \langle (Q_{\factorind})_{i,j}, \: X_i X_j\transpose \right \rangle.
\end{equation}
Equation \eqref{block_decomposition_of_quadratic_factor} and the symmetry of $Q_\factorind$ imply that factor $\ell_\factorind$ depends upon variable $X_\blockvarind$ if and only if there exists some $i \in [\numblockvar]$ such that $(Q_\factorind)_{\blockvarind, i} \ne 0$.  Equivalently:
\begin{equation}
\label{edge_set_in_factor_graph_determines_block_sparsity_pattern_of_objective_data_matrix}
\blockvarind \notin S_\factorind  \Longrightarrow (Q_\factorind)_{\blockvarind, i} = 0 \textnormal{ for all } i \in [\numblockvar].
\end{equation}
That is: if variable $X_\blockvarind$ is \emph{not} adjacent to factor $\ell_\factorind$ in $\Graph$, then the $\blockvarind$-th block row and column of $Q_\factorind$ are both \emph{identically zero}.  Thus, $\Graph$'s edge set $\Edges$ \eqref{edge_set_for_QCQP_factor_graph} determines a specific block sparsity pattern \eqref{edge_set_in_factor_graph_determines_block_sparsity_pattern_of_objective_data_matrix} that the data matrix $Q_\factorind$ parameterizing the $\factorind$-th factor $\ell_\factorind$ [via \eqref{parameterization_of_quadratic_loss_function_in_factor_graph_objective}] must satisfy.

Similarly, let us now consider the data matrices $A_\constraintind$ determining the feasible set in \eqref{QCQP_definition}.  Recall that, by construction, the feasible set for the MLE determined by the factor graph $\Graph$ [cf.\ \eqref{factor_graph_MLE}] is a \emph{Cartesian product}: 
\begin{equation}
\label{cartesian_prod}
\ParamSpace = \ParamSpace_1 \times \dots \times \ParamSpace_\numblockvar;
\end{equation}
i.e., each of the  variables $X_\blockvarind \in \ParamSpace_\blockvarind$ can be varied within its domain $\ParamSpace_\blockvarind$ \emph{independently} of all others.  But this immediately implies that the quadratic constraints in \eqref{QCQP_definition} must be \emph{separable} over the variables $X_1, \dotsc, X_\numblockvar$.\footnote{Note that if any constraint appearing in \eqref{QCQP_definition} involved more than one variable (say $X_i$ and $X_j$ for $i \ne j)$, this would induce a nontrivial coupling between $X_i$ and $X_j$ that would violate their independence  in  \eqref{cartesian_prod}.}   Consequently, we may partition the index set $[\numconstraints]$ for the constraints into $\numblockvar$ subsets $L_1, \dotsc, L_\numblockvar$, where each $L_\blockvarind \subseteq [\numconstraints]$ contains the indices of the constraints in \eqref{QCQP_definition} associated with variable $X_\blockvarind$.  Moreover, given any $\constraintind \in L_\blockvarind$, the argument used to derive \eqref{parameterization_of_quadratic_loss_function_in_factor_graph_objective}--\eqref{edge_set_in_factor_graph_determines_block_sparsity_pattern_of_objective_data_matrix} implies that if  $A_\constraintind$ is partitioned conformally with the block decomposition \eqref{variable_block_decomposition}, then $(A_\constraintind)_{i,j} = 0$ for all $(i,j) \ne (\blockvarind, \blockvarind)$; i.e., $A_\constraintind$ has only a \emph{single nonzero block}, which must appear on the main block diagonal in the $(\blockvarind,\blockvarind)$-th position:
\begin{equation}
\label{block_sparsity_pattern_for_separable_constraint_matrices}
\constraintind \in L_\blockvarind \Longrightarrow A_\constraintind = 
\begin{bmatrix}
0  &        &                &        & \\
   & \ddots &                &        & \\
   &        & (A_{\constraintind})_{\blockvarind, \blockvarind}     &        & \\
   &        &                & \ddots & \\
   &        &                &        & 0 
\end{bmatrix}.
\end{equation}
It follows that the constraints in \eqref{QCQP_definition} can be written more concisely as:
\begin{equation}
\label{eq:partitioned_qcqp_constraints}
    \langle(A_{\constraintind})_{\blockvarind, \blockvarind},X_\blockvarind\,X_\blockvarind^{\top}\rangle=b_{\constraintind},\;\forall \constraintind \in L_\blockvarind,\ \blockvarind\in[\numblockvar].
\end{equation}
In particular, the domain $\ParamSpace_\blockvarind$ for the variable $X_\blockvarind$ is the algebraic variety determined by the constraints indexed by $L_\blockvarind$:
\begin{equation}
\label{block_variable_domain_as_algebraic_variety}
\ParamSpace_\blockvarind = \left \lbrace X_\blockvarind \in \R^{\totalprobdim_\blockvarind \times d} \mid \langle (A_\constraintind)_{\blockvarind, \blockvarind}, \: X_\blockvarind X_\blockvarind\transpose \rangle = b_\constraintind \; \forall \constraintind \in L_\blockvarind \right \rbrace.
\end{equation}

Taken together, these observations show that the structure of the factor graph $\Graph$ defined in \eqref{explicit_factor_graph_model_for_QCQP} implies that the QCQP \eqref{QCQP_definition} must decompose into the following block form:
\begin{equation}
\label{sparse_block_decomposition_of_QCQP}
\begin{split}
&\min_{X_{\blockvarind} \in \R^{\totalprobdim_\blockvarind \times d}} \sum_{\factorind = 1}^{\numfactors}  \overbrace{\sum_{(i,j) \in S_\factorind \times S_\factorind} \langle (Q_\factorind)_{i,j}, \: X_i X_j\transpose \rangle}^{\ell_\factorind(X_{S_\factorind})} \\
&\; \st \langle(A_{\constraintind})_{\blockvarind, \blockvarind},X_\blockvarind\: X_\blockvarind\transpose \rangle=b_{\constraintind},\;\forall \constraintind \in L_\blockvarind,\ \blockvarind\in[\numblockvar].
\end{split}
\end{equation}

\subsubsection{Block sparsity and separability in QCQP data matrices implies factor graph structure} 
\label{block_sparsity_and_separability_implies_factor_graph_structure_subsubsection}

Conversely, consider a QCQP of the form \eqref{QCQP_definition}, and suppose that there exists a row partition \eqref{variable_block_decomposition} of the decision variable $X$ such that \eqref{QCQP_definition} admits a sparse block decomposition of the form \eqref{sparse_block_decomposition_of_QCQP} with respect to the variables $X_1, \dotsc, X_\numblockvar$.  We will show that the bipartite graph $\Graph = (\Variables, \Factors, \Edges)$ defined by \eqref{explicit_factor_graph_model_for_QCQP} is a valid factor graph representation of the QCQP \eqref{sparse_block_decomposition_of_QCQP}.

The argument proceeds by simply reversing the logic of Section \ref{factor_graph_structure_implies_block_sparsity_and_separability_subsubsection}.  Since the constraints in \eqref{sparse_block_decomposition_of_QCQP} are (by construction) separable over $X_1, \dotsc, X_{\numblockvar}$, the feasible set $\ParamSpace$ of \eqref{sparse_block_decomposition_of_QCQP} takes the form of the Cartesian product shown in \eqref{cartesian_prod}, where the domain $\ParamSpace_\blockvarind$ of the $\blockvarind$-th variable $X_\blockvarind$ is given by \eqref{block_variable_domain_as_algebraic_variety}.  Consequently, the set $\Variables$ defined in \eqref{variable_set_for_QCQP_factor_graph} is a valid variable set for a factor graph model of \eqref{sparse_block_decomposition_of_QCQP}.  Similarly, the factor set $\Factors$ defined in \eqref{factor_set_for_QCQP_factor_graph} is (by construction) in one-to-one correspondence with the quadratic summands labeled $\ell_\factorind$ in the objective of \eqref{sparse_block_decomposition_of_QCQP}, as required.  Finally, it is immediate [by inspection of the objective in \eqref{sparse_block_decomposition_of_QCQP}] that factor $\ell_\factorind$ only depends upon those variables $X_\blockvarind$ for which $\blockvarind \in S_\factorind$; consequently, defining the edge set $\Edges$ as in \eqref{edge_set_for_QCQP_factor_graph} ensures that each factor $\ell_\factorind$ is adjacent to every variable $X_\blockvarind$ upon which it explicitly depends.  It follows that the bipartite graph $\Graph = (\Variables, \Factors, \Edges)$ determined by \eqref{explicit_factor_graph_model_for_QCQP} provides a valid factor graph model of the QCQP \eqref{sparse_block_decomposition_of_QCQP}, as claimed.

\subsection{Shor relaxation and Burer-Monteiro factorization inherit factor graph structure}
\label{lifted_factor_graphs_subsection}

In this subsection, we study how the factor graph structure of a QCQP propagates through the certifiable estimation pipeline described in Section \ref{certifiable_estimation_review_section}. Our main result is that the Burer-Monteiro-factored Shor relaxation inherits a factor graph structure from the original QCQP, in which the constituent variables and factors are simple one-to-one algebraic transformations (\emph{lifts}) of those appearing in the original QCQP's factor graph. 

The key observation underpinning our proof is that \emph{the same data matrices $Q$ and $A_\constraintind$ parameterize both the original QCQP \eqref{QCQP_definition} and its Burer-Monteiro-factored Shor relaxation} [cf.\ equations \eqref{QCQP_definition} and \eqref{rank_p_BM_factorization}].  Since the same data matrices parameterize both problems, it follows that any block sparsity and separability structure implied by a factor graph model of the original QCQP is automatically inherited by its associated BM-factored Shor relaxation.  To see this explicitly, suppose that the QCQP \eqref{QCQP_definition} admits the factor graph model $\Graph = (\Variables, \Factors, \Edges)$ defined in \eqref{explicit_factor_graph_model_for_QCQP}.  The results of Section \ref{factor_graph_structure_implies_block_sparsity_and_separability_subsubsection} then imply that the matrices $Q_\factorind$ and $A_\constraintind$ parameterizing this QCQP must  exhibit the block sparsity patterns shown in \eqref{variable_block_decomposition_of_objective_matrix} and \eqref{block_sparsity_pattern_for_separable_constraint_matrices} (respectively), leading to the sparse block decomposition  shown in \eqref{sparse_block_decomposition_of_QCQP}.  Partitioning the rows of the decision variable $Y$ in \eqref{rank_p_BM_factorization} conformally with those of $X$ in \eqref{variable_block_decomposition}:
\begin{equation}
\label{block_row_partition_of_BM_SDP_decision_variable}
Y = 
\begin{bmatrix}
 Y_1 \\
 \vdots \\
 Y_N
\end{bmatrix},
\end{equation}
it follows that the BM-factored Shor relaxation \eqref{rank_p_BM_factorization} admits the analogous sparse block decomposition:
\begin{equation}
\label{sparse_block_decomposition_of_BM_factored_SDP}
\begin{split}
&\min_{Y_{\blockvarind} \in \R^{\totalprobdim_\blockvarind \times \rankrestriction}} \sum_{\factorind = 1}^{\numfactors}  \overbrace{\sum_{(i,j) \in S_\factorind \times S_\factorind} \langle (Q_\factorind)_{i,j}, \: Y_i Y_j\transpose \rangle}^{\triangleq \lifted{\ell}_\factorind(Y_{S_\factorind})} \\
&\; \st \langle(A_{\constraintind})_{\blockvarind, \blockvarind},\: Y_\blockvarind\: Y_\blockvarind\transpose\rangle=b_{\constraintind},\;\forall \constraintind \in L_\blockvarind,\ \blockvarind\in[\numblockvar].
\end{split}
\end{equation}

Finally, since the BM-factored Shor relaxation \eqref{sparse_block_decomposition_of_BM_factored_SDP} is itself a QCQP, we may apply the argument of Section \ref{block_sparsity_and_separability_implies_factor_graph_structure_subsubsection} to construct an associated factor graph $\lifted{\Graph} = (\lifted{\Variables}, \lifted{\Factors}, \lifted{\Edges})$, where:
\begin{subequations}
\label{explicit_factor_graph_model_for_BM_factored_SDP}
\begin{align}
\lifted{\Variables} &\triangleq \lbrace Y_1, \dotsc, Y_\numblockvar \rbrace, \label{variable_set_for_BM_SDP_factor_graph} \\
\lifted{\Factors} &\triangleq \lbrace \lifted{\ell}_1, \dotsc, \lifted{\ell}_\numfactors \rbrace, \label{factor_set_for_BM_SDP_factor_graph} \\
\lifted{\Edges} &\triangleq \lbrace (Y_\blockvarind, \lifted{\ell}_\factorind) \in \lifted{\Variables} \times \lifted{\Factors} \mid \blockvarind \in S_\factorind \rbrace, \label{edge_set_for_BM_SDP_factor_graph}
\end{align}
\end{subequations}
and the domain $\LiftedParamSpace_\blockvarind$ of the $\blockvarind$-th block variable $Y_\blockvarind$ is the algebraic variety defined by:
\begin{equation}
\label{domain_for_lifted_variable}
\LiftedParamSpace_\blockvarind = \left \lbrace Y_\blockvarind \in \R^{\totalprobdim_\blockvarind \times \rankrestriction} \mid \langle (A_\constraintind)_{\blockvarind, \blockvarind}, \: Y_\blockvarind Y_\blockvarind\transpose \rangle = b_\constraintind \; \forall \constraintind \in L_\blockvarind \right \rbrace.
\end{equation}

By construction, the variable and factor sets for $\lifted{\Graph}$ and $\Graph$ are in one-to-one correspondence, and their connectivities \emph{coincide}.  Indeed, comparing the definitions of the feasible sets $\ParamSpace_\blockvarind$ and $\LiftedParamSpace_\blockvarind$ in \eqref{block_variable_domain_as_algebraic_variety} and \eqref{domain_for_lifted_variable} and the factors $\ell_\factorind$ and $\lifted{\ell}_\factorind$ in \eqref{sparse_block_decomposition_of_QCQP} and \eqref{sparse_block_decomposition_of_BM_factored_SDP} (respectively) reveals that the variables and factors appearing in $\lifted{\Graph}$ are obtained from those in $\Graph$ via simple and explicit algebraic transformations (\emph{lifts}).  We refer to these higher-dimensional variable and factor types as \emph{lifted} (or \emph{certifiable}) variables and factors, and the corresponding factor graph $\lifted{\Graph}$ as a \emph{lifted} (or \emph{certifiable}) \emph{factor graph}.  Taken together, these observations establish that the lifted factor graph $\lifted{\Graph}$ \emph{inherits} the structure of the factor graph $\Graph$ from which it is constructed.

The practical significance of the preceding analysis is that if we are given a factor graph model $\Graph$ for a QCQP of the form \eqref{QCQP_definition}, we can obtain a factor graph model $\lifted{\Graph}$ for its associated BM-factored Shor relaxation \eqref{rank_p_BM_factorization} simply by replacing the individual variables and factors appearing in $\Graph$ with their lifted counterparts.  In particular, this means that we can easily implement the low-dimensional \emph{local} optimizations required in the Riemannian Staircase (Alg.\ \ref{Riemannian_Staircase_algorithm}) using current state-of-the-art factor graph-based modeling and optimization libraries, provided that they have been augmented with the requisite certifiable variable and factor types.

\begin{remark}[On block-diagonal constraints]
The preceding discussion highlights two important structural observations regarding block-diagonal constraints. First, the ability to interpret (and instantiate) the BM-factored Shor relaxation \eqref{sparse_block_decomposition_of_BM_factored_SDP} as a factor graph optimization is a consequence of the block-diagonal structure \eqref{block_sparsity_pattern_for_separable_constraint_matrices} of the constraint matrices $A_m$.  Second, this block-diagonal structure is itself induced by the \emph{product structure} \eqref{cartesian_prod} in the original QCQP's feasible set.  While block-diagonal SDP constraints have been extensively studied in the specific context of problems with \emph{orthonormality constraints} \cite{boumal2015riemannian}, the more general origin of such constraints in \emph{product structure}, and the resulting implications for low-rank semidefinite optimization of Shor relaxations, does not appear to have been explicitly recognized in prior work \cite{boumal2015riemannian,boumal2016non,rosen2020Scalable}.  Recognizing this more general structural relationship (linking product structure in the original QCQP to product structure in the BM-factored Shor relaxation) is precisely what enables the certifiable factor graph optimization framework we propose in this paper.
\end{remark}

\subsection{Block separability enables efficient certificate computation }
\label{efficient_certificate_computation_subsection}

In this subsection, we show that the block separability of the constraints in the BM-factored Shor relaxation \eqref{sparse_block_decomposition_of_BM_factored_SDP} implied by the factor graph structure $\lifted{\Graph}$ in \eqref{explicit_factor_graph_model_for_BM_factored_SDP} can also be exploited to improve the computational efficiency of constructing the certificate matrix $S$ \eqref{certificate_matrix_definition} employed in the optimality verification test in the Riemannian Staircase (Algorithm \ref{Riemannian_Staircase_algorithm}).

The key insight enabling this result is that the block separability of the constraints in \eqref{sparse_block_decomposition_of_BM_factored_SDP} induces a corresponding block-diagonal structure in the adjoint operator $\mathcal{A}^*$ defined in \eqref{adjoint_operator_definition}.  Specifically, recalling the partition $L_1, \dotsc, L_\numblockvar \subseteq [\numconstraints]$ of the constraints in \eqref{sparse_block_decomposition_of_BM_factored_SDP} and the induced block-sparsity pattern \eqref{block_sparsity_pattern_for_separable_constraint_matrices} in the corresponding constraint data matrices $A_\constraintind$, we may express the adjoint operator $\mathcal{A}^*$ more explicitly as:
\begin{equation}
\label{block_decomposition_for_adjoint_operator}
\begin{split}
\mathcal{A}^*(\lambda) &= \sum_{\constraintind = 1}^{\numconstraints} \lambda_\constraintind A_\constraintind = \sum_{\blockvarind = 1}^{\numblockvar} \sum_{\constraintind \in L_\blockvarind} \lambda_\constraintind A_\constraintind \\
&= 
\begin{bmatrix}
 \sum_{\constraintind \in L_1} \lambda_\constraintind (A_\constraintind)_{1,1} \\
 & \ddots \\
 & & \sum_{\constraintind \in L_\numblockvar} \lambda_\constraintind (A_\constraintind)_{\numblockvar,\numblockvar}
\end{bmatrix}.
\end{split}
\end{equation}
Equation \eqref{block_decomposition_for_adjoint_operator} reveals that $\mathcal{A}^*(\lambda)$ takes the form of a block-diagonal matrix in which the $\blockvarind$-th diagonal block is constructed using \emph{only} the constraint matrices $A_\constraintind$ and Lagrange multipliers $\lambda_\constraintind$ associated with the $\blockvarind$-th variable $Y_\blockvarind$.  In particular, each of these $\numblockvar$ diagonal blocks can be constructed independently and in parallel, enabling the certificate matrix $S = Q + \mathcal{A}^*(\lambda)$ to be assembled efficiently.

Moreover, the block-diagonal structure of $\mathcal{A}^*(\lambda)$ also enables the efficient parallel computation of the Lagrange multipliers $\lambda \in \R^{\numconstraints}$ themselves.  To see this, note that given a feasible (but not necessarily optimal) point $Y \in \R^{\totalprobdim \times \rankrestriction}$ of the BM-factored Shor relaxation \eqref{rank_p_BM_factorization}, one can calculate a corresponding set of optimal \emph{Lagrange multiplier estimates} $\lambda_{LS}\opt \in \R^{\numconstraints}$ by choosing them to minimize the norm of the stationarity residual in the KKT conditions \eqref{KKT_conditions_for_BM_factorization}:
\begin{equation}
\label{LS_Lagrange_multiplier_estimates_definition}
\begin{split}
\lambda_{LS}\opt &= \argmin_{\lambda \in \R^{\numconstraints}} \left \lVert S(\lambda) Y \right \rVert^2 \\
&= \argmin_{\lambda \in \R^{\numconstraints}} \left \lVert \left[Q + \mathcal{A}^*(\lambda) \right] Y \right \rVert^2.
\end{split}
\end{equation}
The estimates $\lambda_{LS}\opt$ so obtained are referred to as \emph{least squares Lagrange multiplier estimates}.\footnote{Note that these estimates are unique (i.e.\ the right-hand side of \eqref{LS_Lagrange_multiplier_estimates_definition} admits a unique minimizer) whenever the linear independence constraint qualification is satisfied at $Y$; moreover, in that case, the optimal value of the minimization in \eqref{LS_Lagrange_multiplier_estimates_definition} is $0$ if and only if $Y$ is a first-order stationary point of \eqref{rank_p_BM_factorization}, in which case the $\lambda_{LS}\opt$ are the associated Lagrange multipliers \emph{certifying} this fact.}  Recalling the block decompositions \eqref{block_row_partition_of_BM_SDP_decision_variable} and \eqref{block_decomposition_for_adjoint_operator}, we may expand \eqref{LS_Lagrange_multiplier_estimates_definition} as:
\begin{equation}
\label{block_decomposition_of_LS_Lagrange_multiplier_residual}
\begin{split}
\lambda_{LS}\opt &= \argmin_{\lambda \in \R^{\numconstraints}} 
\left \lVert 
\begin{bmatrix}
(QY)_1 \\
\vdots \\
(QY)_\numblockvar
\end{bmatrix}
+ 
\begin{bmatrix}
\sum_{\constraintind \in L_1} \lambda_\constraintind (A_\constraintind)_{1,1} Y_1 \\
\vdots \\
\sum_{\constraintind \in L_\numblockvar} \lambda_\constraintind (A_\constraintind)_{\numblockvar,\numblockvar}
\end{bmatrix}
\right \rVert^2 \\
&= \argmin_{\lambda \in \R^{\numconstraints}} \sum_{\blockvarind = 1}^{\numblockvar} \left \lVert (QY)_\blockvarind + \sum_{\constraintind \in L_\blockvarind} \lambda_\constraintind (A_\constraintind)_{\blockvarind,\blockvarind} Y_\blockvarind \right \rVert^2.
\end{split}
\end{equation}
Note that since the sets $L_\blockvarind$ \emph{partition} the set of constraint indices $[\numconstraints]$ (by construction), the  multipliers $\lambda_{L_\blockvarind}$ associated with the $\blockvarind$-th variable $Y_\blockvarind$ appear \emph{only} in the $\blockvarind$-th summand in the final line of \eqref{block_decomposition_of_LS_Lagrange_multiplier_residual}; that is, the minimization \eqref{block_decomposition_of_LS_Lagrange_multiplier_residual} is \emph{separable} over the  multipliers $\lambda_{L_\blockvarind}$.  Consequently, we can recover these Lagrange multiplier estimates by solving $\numblockvar$ \emph{independent} (and \emph{low-dimensional}) linear least-squares problems in parallel:
\begin{equation}
\label{parallel_minimization_for_least_squares_Lagrange_multiplier_estimates}
\lambda_{L_\blockvarind}^* = \argmin_{\lambda_{L_{\blockvarind}} \in \R^{\lvert L_\blockvarind \rvert}}  \left \lVert (QY)_\blockvarind + \sum_{\constraintind \in L_\blockvarind} \lambda_\constraintind (A_\constraintind)_{\blockvarind,\blockvarind} Y_\blockvarind \right \rVert^2.
\end{equation}

\subsection{The Riemannian Staircase over factor graphs}
\label{subsection:riemannian_staircase_fg}

The results of Sections \ref{factor_graph_structure_in_QCQPs_subsection}--\ref{efficient_certificate_computation_subsection} together enable the Riemannian Staircase to be easily implemented and deployed using the same mature, highly-performant factor graph libraries and associated workflows already ubiquitously employed throughout robotics and computer vision.  This factor graph-based instantiation of the Riemannian Staircase is summarized as Algorithm \ref{Riemannian_Staircase_over_factor_graphs_algorithm}.

\begin{algorithm}[t]
\caption{The Riemannian Staircase over factor graphs}\label{Riemannian_Staircase_over_factor_graphs_algorithm}
\begin{algorithmic}[1]
\Input  Initial feasible point $Y = (Y_1, \dotsc, Y_\numblockvar) \in \R^{\totalprobdim \times \rankrestriction}$ for $\rankrestriction$-dimensional  BM-factored Shor relaxation \eqref{sparse_block_decomposition_of_BM_factored_SDP}. 
\Output Symmetric factor $Y\opt = (Y_1\opt, \dotsc, Y_\numblockvar\opt)$ for minimizer $Z\opt = Y\opt {Y\opt}\transpose$ of Shor relaxation, optimal value $ f_{\mathrm{SDP}}^{*}$.
\Function{RiemannianStaircase}{$Y$}
\Loop 
\Statex[2]{\textcolor{blue}{// Find first-order KKT point of \eqref{sparse_block_decomposition_of_BM_factored_SDP}.}}
\State Construct lifted factor graph model $\lifted{\Graph} = (\lifted{\Variables}, \lifted{\Factors}, \lifted{\Edges})$ 
\Statex[2] for rank-$\rankrestriction$ BM factorization via \eqref{sparse_block_decomposition_of_BM_factored_SDP}--\eqref{domain_for_lifted_variable}.
\State $Y^\star$ $\gets$ \Call{FactorGraphOptimization}{$\lifted{\Graph}, Y$}.  \label{factor_graph_local_opt_step}
\State Compute Lagrange multipliers $\lambda\opt \in \R^{\numconstraints}$ via \eqref{parallel_minimization_for_least_squares_Lagrange_multiplier_estimates}.
\label{parallel_Lagrange_multiplier_computations_step}
\State Construct certificate $S$ from $\lambda^\star$ via \eqref{certificate_matrix_definition} and \eqref{block_decomposition_for_adjoint_operator}. \label{parallel_computation_of_adjoint_diagonal_blocks_step}
\State $(\lambda_{\min}, v_{\min})$ $\gets$ \Call{MinimumEigenpair}{$S$}.
\Statex[2]{\textcolor{blue}{// Test optimality of $Z^\star = Y^\star {Y^\star}\transpose$ for SDP.}}
\If{$\lambda_{\min} \ge 0$} 
\Statex[3]{\textcolor{blue}{// $Z^\star = Y^\star {Y^\star}\transpose$ minimizes SDP by Thm.~\ref{verification_and_saddle_escape_theorem}(a).}}
\State $f_{\mathrm{SDP}}^{*} \gets \langle Q, Y^\star {Y^\star}\transpose\rangle$.
\State \Return $\lbrace Y^\star,  \: f_{\mathrm{SDP}}^{*}\rbrace$
\EndIf
\Statex[2]{\textcolor{blue}{// Construct lifted embedding of $Y^\star$.}}
\State $Y_{+} \gets \left[Y^\star \mid 0\right]$
\Statex[2]{\textcolor{blue}{// Construct second-order descent direction.}}
\State $\dot{Y}_{+} \gets \left[0 \mid v_{\min}\right]$
\Statex[2]{\textcolor{blue}{// Backtracking line search for saddle escape.}}
\State $Y$ $\gets$ \Call{LineSearch}{$Y_{+}, \dot{Y}_{+}$}
\EndLoop
\EndFunction
\end{algorithmic}
\end{algorithm}

Note that Algorithm \ref{Riemannian_Staircase_over_factor_graphs_algorithm} is simply a concrete realization of the generic Riemannian Staircase (Algorithm \ref{Riemannian_Staircase_algorithm}) that makes explicit how several of the required (abstract) operations (namely local optimization, Lagrange multiplier computation, and certificate matrix construction) can be efficiently implemented within the factor graph paradigm.   Specifically, in Algorithm \ref{Riemannian_Staircase_over_factor_graphs_algorithm} the local optimization required at each level of the Riemannian Staircase (line \ref{BM_local_opt_step} of Algorithm \ref{Riemannian_Staircase_algorithm}) is carried out in line \ref{factor_graph_local_opt_step} as a standard factor graph optimization over the rank-$\rankrestriction$  lifted factor graph $\lifted{\Graph} = (\lifted{\Variables}, \lifted{\Factors}, \lifted{\Edges})$, which can be automatically synthesized directly from the factor graph model ${\Graph} = ({\Variables}, {\Factors}, {\Edges})$ specifying the original QCQP estimation task  using the constructive lifting (i.e.\ one-to-one variable and factor substitution) procedure described in Section \ref{lifted_factor_graphs_subsection}.  Given the first-order stationary point $Y^*$ recovered from this factor graph optimization, an associated set of optimal Lagrange multipliers $\lambda^*$ can then be computed efficiently (in line \ref{parallel_Lagrange_multiplier_computations_step}) via the $\numblockvar$ independent low-dimensional linear least-squares solves described in Section \ref{efficient_certificate_computation_subsection}.\footnote{This efficient computation of Lagrange multiplier estimates is particularly important when the factor graph optimizations in line \ref{factor_graph_local_opt_step} are implemented using manifold-based optimization algorithms (such as those employed in e.g.\ GTSAM \cite{Dellaert2012GTSAM} or Manopt \cite{Boumal2014Manopt}): since these methods are \emph{intrinsic} (i.e., they formulate the problem as \emph{unconstrained} optimization over a smooth manifold), they do not natively return an associated set of Lagrange multipliers.}  Finally, the $\numblockvar$ diagonal blocks of the adjoint operator $\mathcal{A}^*(\lambda^*)$ required to construct the certificate matrix $S$ in \eqref{certificate_matrix_definition} can then be assembled independently and in parallel (in line \ref{parallel_computation_of_adjoint_diagonal_blocks_step}) as shown in \eqref{block_decomposition_for_adjoint_operator}.  Crucially, both of the latter two operations (Lagrange multiplier computation and adjoint operator evaluation) are \emph{separable} over the individual variables $Y_\blockvarind$, and the contribution associated with each variable $Y_\blockvarind$ depends only upon its own value and the quadratic equations defining its domain $\LiftedParamSpace_\blockvarind$ in \eqref{domain_for_lifted_variable}.  This separability is in keeping with the modular design philosophy of the factor graph paradigm; in particular, at the level of software, these computations can be implemented as \emph{member functions} within the classes defining the domains $\LiftedParamSpace_\blockvarind$ of the lifted variables $Y_\blockvarind$ appearing in the lifted factor graph $\lifted{G}$.

Finally, we note that the synthesis of the factor graph and certifiable estimation paradigms developed in this section (and instantiated in Algorithm \ref{Riemannian_Staircase_over_factor_graphs_algorithm}) is mutually beneficial.  From the perspective of certifiable estimation, our synthesis dramatically lowers the practical barriers to designing and deploying certifiable estimators: practitioners can now obtain verifiably globally optimal solutions using existing factor graph libraries and workflows, with no specialized expertise in convex optimization or semidefinite programming and no custom solver development required. In short, certifiable estimation becomes as straightforward to implement and deploy as conventional local factor graph optimization.  From the perspective of factor graph inference, the Riemannian Staircase (Algorithm \ref{Riemannian_Staircase_over_factor_graphs_algorithm}) can be viewed as a lightweight meta-algorithm that wraps around standard \emph{local} factor graph optimization, yet is capable of guaranteeing the recovery of \emph{certifiably globally optimal solutions}.  Our synthesis thus achieves the best of both worlds: it combines the robustness and formal global optimality guarantees of certifiable estimation with the simplicity, ease of use, and high computational performance of current state-of-the-art factor graph-based modeling and optimization libraries.

\section{Common examples of lifted variables and factors for certifiable factor graph optimization}
\label{common_lifted_types_section}

In this section, we present the \emph{lifted} analogues of several variable and factor types commonly encountered in robotics and computer vision.  These data types provide the concrete building blocks necessary to implement the experimental evaluations reported in Section \ref{Experimental_results_section}.

\subsection{Common lifted variable types}\label{subsection:Lifting-Variables}

\subsubsection{Rotations and orthogonal matrices}

Following \cite{rosen2019se}, we model both rotations $R \in \SO(d)$ and orthogonal matrices $R  \in \Orthogonal(d)$ as elements of the orthogonal group by enforcing the orthonormality constraints:
\begin{equation}
\label{orthonormality_constraints}
R\transpose R = I_d.
\end{equation}
Following Section \ref{lifted_factor_graphs_subsection}, the lifted variable $Y \in \R^{\rankrestriction \times d}$ corresponding to $R \in \R^{d \times d}$ satisfies the constraints [cf.\ \eqref{domain_for_lifted_variable}]:
\begin{equation}
Y\transpose Y = I_d,
\end{equation}
which we may recognize as the defining equations for the Stiefel manifold $\Stiefel(d, \rankrestriction)$.

\subsubsection{Unit vectors} 
We model unit vectors $v \in S^{d-1}$ in $d$-dimensional Euclidean space with the constraint:
\begin{equation}
v\transpose v = 1.
\end{equation}
The lifted variable $u \in \R^\rankrestriction$ corresponding to $v \in \R^d$ likewise satisfies the constraint:
\begin{equation}
u\transpose u = 1,
\end{equation}
which we recognize as the defining equation for the (higher-dimensional) unit sphere $S^{\rankrestriction - 1}$ in $\rankrestriction$-dimensional Euclidean space.

\subsubsection{Translations}

Translation variables $t \in \R^d$ are \emph{unconstrained}, i.e., the set of quadratic equations appearing on the right-hand side of \eqref{domain_for_lifted_variable} is empty.  Consequently, a $d$-dimensional translation $t$ simply lifts to the higher-dimensional translation $u \in \R^{\rankrestriction}$.

\subsection{Common lifted factor types}


\subsubsection{Relative rotation measurements}

Given two rotations $R_i, R_j \in \SO(d)$, the \emph{relative rotation} $R_{ij} \in \SO(d)$ from $R_i$ to $R_j$ is:
\begin{equation}\label{relative_rotation_definition}
R_{ij} \triangleq R_i^{-1} R_j.
\end{equation}
Following \cite{rosen2019se}, we assume that noisy relative rotation measurements $\tilde{R}_{ij}  \in \SO(d)$ are sampled according to:
\begin{equation}\label{isotropic_Langevin_noise_model}
\tilde{R}_{ij} = R_{ij} \eta_{ij}, \quad \eta_{ij} \sim \operatorname{Langevin}(I_d, \kappa_{ij}),
\end{equation}
where $\operatorname{Langevin}(I_d, \kappa_{ij})$ denotes the \emph{isotropic Langevin distribution} over $\SO(d)$ with concentration parameter  $\kappa_{ij} \geq 0$. The negative log-likelihood function corresponding to the measurement model \eqref{relative_rotation_definition}--\eqref{isotropic_Langevin_noise_model} is then:
\begin{equation}
\label{relative_rotation_factor}
\ell(R_i, R_j; \tilde{R}_{ij}) = \kappa_{ij} \|R_j - R_i \tilde{R}_{ij}\|_F^2.
\end{equation}
Following Section \ref{lifted_factor_graphs_subsection}, the \emph{lifted} relative rotation factor corresponding to \eqref{relative_rotation_factor} is:
\begin{equation}
\label{lifted_relative_rotation_factor}
\lifted{\ell}(Y_i, Y_j; \tilde{R}_{ij}) = \kappa_{ij} \|Y_j - Y_i \tilde{R}_{ij}\|_F^2.
\end{equation}

\subsubsection{Relative translation measurements}

Given a point $t_j \in \R^d$ and a pose $x_i = (t_i, R_i) \in \SE(d)$, the \emph{relative translation} $t_{ij} \in \R^d$ from $x_i$ to $t_j$ (as measured from the coordinate frame defined by $x_i$ itself) is:
\begin{equation}
\label{relative_translation_measurement}
t_{ij} \triangleq R_i\inv(t_j - t_i).
\end{equation}
We assume that noisy relative translation measurements $\tilde{t}_{ij} \in \R^d$ are sampled according to:
\begin{equation}
\label{relative_translation_noise_model}
\tilde{t}_{ij} = t_{ij} + \epsilon_{ij}, \quad \epsilon_{ij} \sim \mathcal{N}(0, \tau_{ij}^{-1} I_d),
\end{equation}
where $\tau_{ij} \geq 0$ denotes the measurement precision. The negative log-likelihood function corresponding to the measurement model \eqref{relative_translation_measurement}--\eqref{relative_translation_noise_model} is then:
\begin{equation}
\label{relative_translation_factor}
\ell(R_i, t_i, t_j; \tilde{t}_{ij}) = \tau_{ij} \|t_j - t_i - R_i \tilde{t}_{ij}\|_2^2.
\end{equation}
Following Section \ref{lifted_factor_graphs_subsection}, the \emph{lifted} relative translation factor corresponding to \eqref{relative_translation_factor} is:
\begin{equation}
\label{lifted_relative_translation_factor}
\lifted{\ell}(Y_i, u_i, u_j; \tilde{t}_{ij}) = \tau_{ij} \|u_j - u_i  - Y_i \tilde{t}_{ij}\|_2^2.
\end{equation}

\subsubsection{Range measurements}
\label{subsubsection:Range_Measurement}

Given two points $t_i, t_j \in \mathbb{R}^d$, the \emph{range} $r_{ij} \in \R_{+}$ between them is:
\begin{equation}
\label{range_measurement}
r_{ij} \triangleq \|t_i - t_j\|_2.
\end{equation}
We assume that noisy range measurements are sampled according to:
\begin{equation}
\label{range_noise_model}
\tilde{r}_{ij} = r_{ij} + \nu_{ij}, \quad \nu_{ij} \sim \mathcal{N}(0, \sigma_{ij}^2),
\end{equation}
where $\sigma_{ij}^2$ is the measurement variance. The negative log-likelihood function corresponding to \eqref{range_measurement}--\eqref{range_noise_model} is then:
\begin{equation}\label{equation: non_lifted_range_factor}
\ell(t_i, t_j; \tilde{r}_{ij}) = \frac{1}{\sigma_{ij}^2} \left(\|t_j - t_i\|_2 - \tilde{r}_{ij}\right)^2.
\end{equation}

Note that~\eqref{equation: non_lifted_range_factor} is \emph{not} a quadratic function of $t_i$ and $t_j$, due to the unsquared norm appearing on the right-hand side. Following~\cite{halsted2022riemannian, papalia2024certifiably}, we introduce an auxiliary variable $b_{ij} \in S^{d-1}$ associated with this measurement that represents the \emph{bearing} from $t_i$ to $t_j$, which can be shown to yield an \emph{equivalent} quadratic formulation:
\begin{equation}\label{quadratic_range_factor}
\ell(t_i, t_j, b_{ij}; \tilde{r}_{ij}) = \frac{1}{\sigma_{ij}^2} \|t_j - t_i - \tilde{r}_{ij} b_{ij}\|_2^2.
\end{equation}
Following Section \ref{lifted_factor_graphs_subsection}, the \emph{lifted} range factor corresponding to \eqref{quadratic_range_factor} is:
\begin{equation}
\label{lifted_relative_translation_factor}
\lifted{\ell}(u_i, u_j, s_{ij}; \tilde{r}_{ij}) = \frac{1}{\sigma_{ij}^2} \|u_j - u_i  - \tilde{r}_{ij} s_{ij} \|_2^2.
\end{equation}

\subsection{Summary}

For ease of reference, we summarize the lifted variable and factor types described in this section in Table~\ref{tab:lifted_summary}.  These lifted types can be composed to formulate a wide variety of certifiable estimators, as demonstrated in our experimental evaluations (Section~\ref{Experimental_results_section}).

\begin{table}[t]
\centering
\caption{Summary of common lifted variable and factor types}
\label{tab:lifted_summary}
\renewcommand{\arraystretch}{1.5}
\begin{tabular}{lll}
\hline
\textbf{Type} & \textbf{Original} & \textbf{Lifted (rank-$\rankrestriction$)} \\
\hline
Rotation & $R \in O(d)$ & $Y \in \Stiefel(d, \rankrestriction)$ \\
Unit sphere & $v \in S^{d-1}$ & $u \in S^{\rankrestriction-1}$ \\
Translation & $t \in \R^d$ & $u \in \R^\rankrestriction$ \\
\hline
Relative rotation & $\kappa_{ij} \|R_j - R_i \tilde{R}_{ij}\|_F^2$ & $\kappa_{ij} \|Y_j - Y_i \tilde{R}_{ij}\|_F^2$ \\
Relative translation & $\tau_{ij} \|t_j - t_i - R_i \tilde{t}_{ij}\|_2^2$ & $\tau_{ij} \|u_j - u_i - Y_i \tilde{t}_{ij}\|_2^2$ \\
Range & $\sigma_{ij}^{-2} \|t_j - t_i - \tilde{r} b\|_2^2$ & $\sigma_{ij}^{-2} \|u_j - u_i - \tilde{r} s\|_2^2$ \\
\hline
\end{tabular}
\end{table}

\newcommand{\NA}{\textemdash}
\newcommand{\tabsize}{\scriptsize}      
\newcommand{\tabstyle}{\tabsize\setlength{\tabcolsep}{3.8pt}\renewcommand{\arraystretch}{1.05}}

\begin{table*}[t]
\centering
\caption{\textsc{\textbf{Pose graph optimization datasets}}}
\tabstyle
\label{Table: PGO_dataset_information}
\begin{tabular}{l *{15}{c}}
\toprule
& MIT & CSAIL & Intel & Manhattan & City
& Small Grid & Garage & Sphere & Torus & Cubicle & Grid & Rim \\
\midrule
\textbf{Dimension}             & 2D & 2D & 2D & 2D & 2D & 3D & 3D & 3D & 3D & 3D & 3D & 3D \\
\textbf{\# Poses}        & 808 & 1045 & 1728 & 3500 & 10000 & 125 & 1661 & 2500 & 5000 & 5750 & 8000 & 10093 \\
\textbf{\# Measurements} & 827 & 1171 & 2827 & 5451 & 20687 & 297 & 6275 & 4949 & 10000 & 7696 & 22236 & 18637 \\
\bottomrule
\end{tabular}
\end{table*}

\begin{table*}[t]
\centering
\caption{\textsc{\textbf{Landmark and range-aided SLAM datasets}}}
\tabstyle
\label{Table: landmark_range_dataset_information}
\begin{tabular}{l *{10}{c}}
\toprule
& \multicolumn{2}{c}{\textbf{Landmark SLAM}} & \multicolumn{8}{c}{\textbf{Range-aided SLAM}} \\
\cmidrule(lr){2-3}\cmidrule(lr){4-11}
& Victoria \cite{guivant2002optimization} & Trees \cite{kaess2012isam2} & Goats 16 \cite{olson2007robust} & Goats 15  & Plaza 1 \cite{djugash2009navigating} & Plaza 2  & MRCLAM 2 \cite{leung2011utias} & MRCLAM 4 & MRCLAM 6 & MRCLAM 7  \\
\midrule
\textbf{Dimension}             & 2D & 2D & 2D & 2D & 2D & 2D & 2D & 2D & 2D & 2D \\
\textbf{\# Poses}        & 6969 & 10000 & 201 & 472 & 9658 & 4091 & 31067 & 26842 & 11274 & 13720 \\
\textbf{\# Landmarks}    & 151 & 100 & 4 & 4 & 4 & 4 & 15 & 15 & 15 & 15 \\
\textbf{\# Measurements} & 10608 & 14442 & 772 & 1258 & 12983 & 5897 & 44857 & 39064 & 16658 & 20060  \\
\bottomrule
\end{tabular}
\end{table*}

\providecommand{\NA}{\textemdash}
\providecommand{\ObjOne}[1]{\multicolumn{2}{c}{#1}}

\begin{table*}[t]
\centering
\caption{\textsc{\textbf{Comparison with specialized certifiable estimators}}}
\tabstyle
\label{Table: random_initialization_experiments}
\setlength{\tabcolsep}{2.0pt}
{\small
\resizebox{\textwidth}{!}{%
\begin{tabular}{l *{10}{c}}
\toprule
\multicolumn{1}{l}{\textbf{PGO}} & \multicolumn{5}{c}{\textbf{SE-Sync}} & \multicolumn{5}{c}{\textbf{Ours}} \\
\cmidrule(lr){2-6}\cmidrule(lr){7-11}
 & \multicolumn{2}{c}{Obj.~value} & Opt.~time [s] & Tot.~time [s] & Term.~level
 & \multicolumn{2}{c}{Obj.~value} & Opt.~time [s] & Tot.~time [s] & Term.~level \\
\midrule
MIT
  & \ObjOne{$6.115{\times}10^{1}$} & 0.21 & 0.22 & 4
  & \ObjOne{$6.115{\times}10^{1}$} & 1.09 & 1.15 & 4 \\
CSAIL
  & \ObjOne{$3.170{\times}10^{1}$} & 0.12 & 0.13 & 4
  & \ObjOne{$3.170{\times}10^{1}$} & 1.06 & 1.18 & 4 \\
Intel
  & \ObjOne{$5.235{\times}10^{1}$} & 1.80 & 1.83 & 4
  & \ObjOne{$5.235{\times}10^{1}$} & 3.45 & 3.61 & 4 \\
Manhattan
  & \ObjOne{$2.049{\times}10^{2}$} & 9.24 & 9.36 & 4
  & \ObjOne{$2.049{\times}10^{2}$} & 9.39 & 9.85 & 4 \\
City10000
  & \ObjOne{$6.386{\times}10^{2}$} & 45.77 & 46.12 & 4
  & \ObjOne{$6.386{\times}10^{2}$} & 23.54 & 24.93 & 4 \\
Small Grid
  & \ObjOne{$1.025{\times}10^{3}$} & 0.07 & 0.08 & 4
  & \ObjOne{$1.025{\times}10^{3}$} & 0.06 & 0.06 & 3 \\
Garage
  & \ObjOne{$1.264{\times}10^{0}$} & 37.41 & 37.52 & 5
  & \ObjOne{$1.263{\times}10^{0}$} & 11.39 & 11.75 & 5 \\
Sphere
  & \ObjOne{$1.687{\times}10^{3}$} & 7.80 & 7.89 & 5
  & \ObjOne{$1.687{\times}10^{3}$} & 7.21 & 7.53 & 5 \\
Torus
  & \ObjOne{$2.423{\times}10^{4}$} & 7.70 & 7.87 & 5
  & \ObjOne{$2.423{\times}10^{4}$} & 90.46 & 91.09 & 5 \\
Cubicle
  & \ObjOne{$7.171{\times}10^{2}$} & 31.31 & 31.45 & 4
  & \ObjOne{$7.171{\times}10^{2}$} & 29.31 & 30.67 & 5 \\
Grid3D
  & \ObjOne{$8.432{\times}10^{4}$} & 40.59 & 42.77 & 5
  & \ObjOne{$8.432{\times}10^{4}$} & 185.98 & 187.63 & 3 \\
Rim
  & \ObjOne{$5.461{\times}10^{3}$} & 153.53 & 154.22 & 5
  & \ObjOne{$5.461{\times}10^{3}$} & 84.20 & 87.74 & 5 \\
\midrule\midrule
\multicolumn{1}{l}{\textbf{Landmark SLAM}} & \multicolumn{5}{c}{\textbf{Landmark SE-Sync}} & \multicolumn{5}{c}{\textbf{Ours}} \\
\cmidrule(lr){2-6}\cmidrule(lr){7-11}
& \multicolumn{2}{c}{Obj.~value} & Opt.~time [s] & Tot.~time [s] & Term.~level
& \multicolumn{2}{c}{Obj.~value} & Opt.~time [s] & Tot.~time [s] & Term.~level \\
\midrule
Victoria
  & \ObjOne{$4.660{\times}10^{2}$} & 81.17 & 81.96 & 4
  & \ObjOne{$4.680{\times}10^{2}$} & 21.95 & 23.14 & 4 \\
Trees
  & \ObjOne{$6.035{\times}10^{2}$} & 64.93 & 65.59 & 4
  & \ObjOne{$6.035{\times}10^{2}$} & 18.45 & 20.26 & 4 \\
\midrule\midrule
\multicolumn{1}{l}{\textbf{Range-aided SLAM}} & \multicolumn{5}{c}{\textbf{CORA}} & \multicolumn{5}{c}{\textbf{Ours}} \\
\cmidrule(lr){2-6}\cmidrule(lr){7-11}
& SDP~value & Ref.~value & Opt.~time [s] & Tot.~time [s] & Term.~level
& SDP~value & Ref.~value & Opt.~time [s] & Tot.~time [s] & Term.~level \\
\midrule
Goats 15
  & $1.607{\times}10^{4}$ & $1.820{\times}10^{4}$ & 0.96 & 0.99 & 5
  & $1.614{\times}10^{4}$ & $1.820{\times}10^{4}$ & 1.28 & 1.40 & 4 \\
Goats 16
  & $3.686{\times}10^{3}$ & $3.894{\times}10^{3}$ & 0.49 & 0.51 & 5
  & $3.718{\times}10^{3}$ & $3.894{\times}10^{3}$ & 1.41 & 1.51 & 4 \\
Plaza 1
  & $1.280{\times}10^{3}$ & $1.333{\times}10^{3}$ & 10.81 & 11.29 & 4
  & $1.291{\times}10^{3}$ & $1.333{\times}10^{3}$ & 125.71 & 141.40 & 6 \\
Plaza 2
  & $7.243{\times}10^{2}$ & $7.343{\times}10^{2}$ & 2.03 & 2.14 & 3
  & $7.302{\times}10^{2}$ & $7.343{\times}10^{2}$ & 30.21 & 32.01 & 4 \\
MRCLAM 2
  & $7.033{\times}10^{3}$ & $7.040{\times}10^{3}$ & 51.03 & 53.32 & 4
  & $7.043{\times}10^{3}$ & $7.040{\times}10^{3}$ & 63.68 & 89.29 & 4 \\
MRCLAM 4
  & $4.173{\times}10^{3}$ & $4.173{\times}10^{3}$ & 33.52 & 35.64 & 4
  & $4.173{\times}10^{3}$ & $4.173{\times}10^{3}$ & 61.87 & 80.62 & 4 \\
MRCLAM 6
  & $3.124{\times}10^{3}$ & $3.144{\times}10^{3}$ & 23.32 & 24.13 & 4
  & $3.291{\times}10^{3}$ & $3.144{\times}10^{3}$ & 22.77 & 32.10 & 4 \\
MRCLAM 7
  & $3.040{\times}10^{3}$ & $3.110{\times}10^{3}$ & 14.11 & 14.61 & 3
  & $3.044{\times}10^{3}$ & $3.110{\times}10^{3}$ & 25.16 & 34.45 & 4 \\
\bottomrule
\end{tabular}
}
}
\captionsetup{justification=justified,singlelinecheck=false}
\caption*{\footnotesize
This table provides an empirical comparison of our certifiable factor graph-based approach with state-of-the-art \emph{specialized}, \emph{problem-specific} certifiable estimators (specifically SE-Sync \cite{rosen2019se}, Landmark SE-Sync \cite{fan2020cpl}, and CORA \cite{papalia2024certifiably}) on several large-scale benchmarks for pose graph optimization, landmark SLAM, and range-aided SLAM.  Here \emph{Opt.~time} denotes the cumulative time spent performing local optimization across all levels of the Riemannian Staircase, \emph{Tot.~time} denotes the total runtime (including local optimization, verification, saddle escape, and rounding), and \emph{Term.~level} denotes the terminal level of the Riemannian Staircase.   For PGO and landmark SLAM problems, the SDP relaxation is exact for all test cases, so \emph{Obj.~value} reports the optimal value of \emph{both} the QCQP \eqref{QCQP_definition} and its SDP relaxation \eqref{Shor_relaxation}.  For range-aided SLAM, where the SDP relaxation is generally inexact, we separately report both the optimal value of the SDP relaxation  (\emph{SDP~value}) and the value of the \emph{refined} estimate (\emph{Ref.~value}) obtained by locally optimizing the rounded feasible estimate recovered from the SDP (cf.\ Section \ref{certifiable_estimator_comparison_implementation_details_subsection}).
}
\vspace{-\baselineskip}
\end{table*}

\begin{table*}[t]
\centering
\caption{\textsc{\textbf{Comparison with local factor graph-based optimization}}}
\tabstyle
\label{Table: odom_initialization_experiments}
\setlength{\tabcolsep}{2.0pt}
\providecommand{\NA}{\textemdash}
\providecommand{\ObjOne}[1]{\multicolumn{2}{c}{#1}}

\resizebox{\textwidth}{!}{%
{\small
\begin{tabular}{l *{10}{c}}
\toprule
\multicolumn{1}{l}{\textbf{PGO}} & \multicolumn{7}{c}{\textbf{Ours}} & \multicolumn{3}{c}{\textbf{LM (GTSAM)}} \\
\cmidrule(lr){2-8}\cmidrule(lr){9-11}
& & \multicolumn{2}{c}{Obj.~value} & Opt.~time [s] & Tot.~time [s] & Init.~level & Term.~level 
 & Obj.~value & Opt.~time [s] & Verif. \\
\midrule
MIT
&  & \ObjOne{$6.115{\times}10^{1}$} & 0.40 & 0.47 & 2 & 4
  & $1.298{\times}10^{3}$ & 0.12 & \ding{55} \\
CSAIL
&  & \ObjOne{$3.170{\times}10^{1}$} & 0.04 & 0.05 & 2 & 2 
  & $3.170{\times}10^{1}$ & 0.05 & \checkmark \\
Intel
&  & \ObjOne{$5.235{\times}10^{1}$} & 0.05 & 0.07 & 2 & 2 
  & $5.235{\times}10^{1}$ & 0.07 & \checkmark \\
Manhattan
&  & \ObjOne{$2.049{\times}10^{2}$} & 0.52 & 0.57 & 2 & 2 
  & $2.049{\times}10^{2}$ & 0.55 & \checkmark \\
City10000
&  & \ObjOne{$6.386{\times}10^{2}$} & 0.36 & 0.56 & 2 & 2 
  & $6.386{\times}10^{2}$ & 0.49 & \checkmark \\
Small Grid
&  & \ObjOne{$1.025{\times}10^{3}$} & 0.01 & 0.02 & 3 & 3 
  & $1.025{\times}10^{3}$ & 0.02 & \checkmark \\
Garage
&  & \ObjOne{$1.263{\times}10^{0}$} & 0.21 & 0.25 & 3 & 3 
  & $1.263{\times}10^{0}$ & 0.23 & \checkmark \\
Sphere
&  & \ObjOne{$1.687{\times}10^{3}$} & 0.18 & 0.24 & 3 & 3 
  & $1.687{\times}10^{3}$ & 0.22 & \checkmark \\
Torus
&  & \ObjOne{$2.423{\times}10^{4}$} & 22.73 & 23.55 & 3 & 5 
  & $5.276{\times}10^{4}$ & 3.93 & \ding{55} \\
Cubicle
&  & \ObjOne{$7.171{\times}10^{2}$} & 13.95 & 15.34 & 3 & 5
  & $2.380{\times}10^{3}$ & 3.86 & \ding{55} \\
Grid3D
&  & \ObjOne{$8.432{\times}10^{4}$} & 61.50 & 63.18 & 3 & 3 
  & $8.432{\times}10^{4}$ & 61.38 & \checkmark \\
Rim
&  & \ObjOne{$5.461{\times}10^{3}$} & 77.93 & 81.48 & 3 & 5
  & $1.510{\times}10^{4}$ & 6.46 & \ding{55} \\
\midrule\midrule
\multicolumn{1}{l}{\textbf{Landmark SLAM}} & \multicolumn{7}{c}{\textbf{Ours}} & \multicolumn{3}{c}{\textbf{LM (GTSAM)}} \\
\cmidrule(lr){2-8}\cmidrule(lr){9-11}
& & \multicolumn{2}{c}{Obj.~value} & Opt.~time [s] & Tot.~time [s] & Init.~level & Term.~level 
& Obj.~value & Opt.~time [s] & Verif. \\
\midrule
Victoria
&  & \ObjOne{$4.661{\times}10^{2}$} & 9.50 & 11.48 & 2 & 4 
  & $1.691{\times}10^{4}$ & 3.27 & \ding{55} \\
Trees
&  & \ObjOne{$6.035{\times}10^{2}$} & 0.28 & 0.42 & 2 & 2 
  & $6.035{\times}10^{2}$ & 0.39 & \checkmark \\
\midrule\midrule

\multicolumn{1}{l}{\textbf{Range-Aided SLAM}} & & \multicolumn{6}{c}{\textbf{Ours}}  & \multicolumn{3}{c}{\textbf{LM (GTSAM)}} \\
\cmidrule(lr){3-8}\cmidrule(lr){9-11}
& & SDP~value & Ref.~value & Opt.~time [s] & Tot.~time [s] & Init.~level & Term.~level 
& \multicolumn{2}{c}{Obj.~value} & Opt.~time [s] \\
\midrule
Goats 15 &
  & $1.739{\times}10^{4}$ & $1.820{\times}10^{4}$ & 3.62 & 4.09 & 2 & 5
  & \multicolumn{2}{c}{$1.820{\times}10^{4}$} & 0.06  \\
Goats 16 &
  & $4.106{\times}10^{3}$ & $3.894{\times}10^{3}$ & 0.67 & 0.79 & 2 & 4 
  & \multicolumn{2}{c}{$2.713{\times}10^{4}$} & 0.03  \\
Plaza 1 &
  & $1.295{\times}10^{3}$ & $1.333{\times}10^{3}$ & 82.03 & 88.47 & 2 & 4
  & \multicolumn{2}{c}{$4.037{\times}10^{5}$} & 3.56  \\
Plaza 2 &
  & $7.776{\times}10^{2}$ & $7.343{\times}10^{2}$ & 29.22 & 31.06 & 2 & 4
  & \multicolumn{2}{c}{$2.182{\times}10^{4}$} & 1.75 \\
MRCLAM 2 &
  & $7.034{\times}10^{3}$ & $7.041{\times}10^{3}$ & 60.79 & 89.59 & 2 & 4
  & \multicolumn{2}{c}{$1.330{\times}10^{4}$} & 14.51  \\
MRCLAM 4 &
  & $4.173{\times}10^{3}$ & $4.173{\times}10^{3}$ & 51.75 & 79.26 & 2 & 4
  & \multicolumn{2}{c}{$2.046{\times}10^{4}$} & 32.00  \\
MRCLAM 6 &
  & $3.132{\times}10^{3}$ & $3.144{\times}10^{3}$ & 22.50 & 26.89 & 2 & 3
  & \multicolumn{2}{c}{$3.619{\times}10^{3}$} & 6.09  \\
MRCLAM 7 &
  & $3.104{\times}10^{3}$ & $3.110{\times}10^{3}$ & 19.54 & 24.88 & 2 & 3
  & \multicolumn{2}{c}{$4.099{\times}10^{3}$} & 6.03  \\
\bottomrule
\end{tabular}
}%
} 
\captionsetup{justification=justified,singlelinecheck=false}
\caption*{\footnotesize
This table provides an empirical comparison of our certifiable factor graph-based approach with a state-of-the-art \emph{local} factor graph-based optimization method, namely the Levenberg–Marquardt (LM) solver provided by the GTSAM library \cite{gtsam}, on several large-scale benchmarks for pose graph optimization, landmark SLAM, and range-aided SLAM. For our method, \emph{Opt.~time} denotes the cumulative time spent performing local optimizations across all levels of the Riemannian Staircase, while \emph{Tot.~time} includes all components of the algorithm (local optimization, certification, saddle escape, and rounding). The columns \emph{Init.~level} and \emph{Term.~level} report the initial and terminal levels of the Riemannian Staircase, respectively. For the local LM solver, \emph{Verif.} indicates whether the solution returned by local optimization is certifiably globally optimal (\checkmark) or not (\ding{55}); this column is shown only for PGO and landmark SLAM problems, where the SDP relaxation is exact and a posterior optimality certificate is therefore applicable. For PGO and landmark SLAM problems, the SDP relaxation is exact for all test cases, so \emph{Obj.~value} reports the optimal value of both the QCQP and its SDP relaxation, while for range-aided SLAM instances we separately report the SDP relaxation value (\emph{SDP~value}) and the objective value of the refined estimate (\emph{Ref.~value}) obtained after local optimization of the rounded solution.
}
\end{table*}

\section{Experiments}
\label{Experimental_results_section}

In this section, we evaluate the certifiable factor graph optimization framework developed in Section \ref{Certifiable_factor_graph_optimization_section} on standard benchmarks for three representative SLAM problem classes: pose graph optimization, landmark SLAM, and range-aided SLAM. Our evaluation is designed to assess the performance of our approach relative to the two estimator paradigms that it unifies: specifically, we compare against state-of-the-art certifiable estimators to verify our approach's functional equivalence with \emph{specialized}, \emph{problem-specific} methods, and against standard \emph{local} factor graph-based optimization methods to evaluate whether our approach preserves their high computational efficiency while additionally enabling the recovery of \emph{certifiably optimal or near-optimal} solutions.


%

\begin{figure*}[htbp]
  \centering

  \makebox[0.3\textwidth]{\centering \textbf{Pose graph optimization}}%
  \hfill
  \makebox[0.3\textwidth]{\centering \textbf{Landmark SLAM}}%
  \hfill
  \makebox[0.3\textwidth]{\centering \textbf{Range-aided SLAM}}%

  \vspace{0.5em}

  \begin{subfigure}[b]{0.3\textwidth}
    \includegraphics[width=\textwidth]{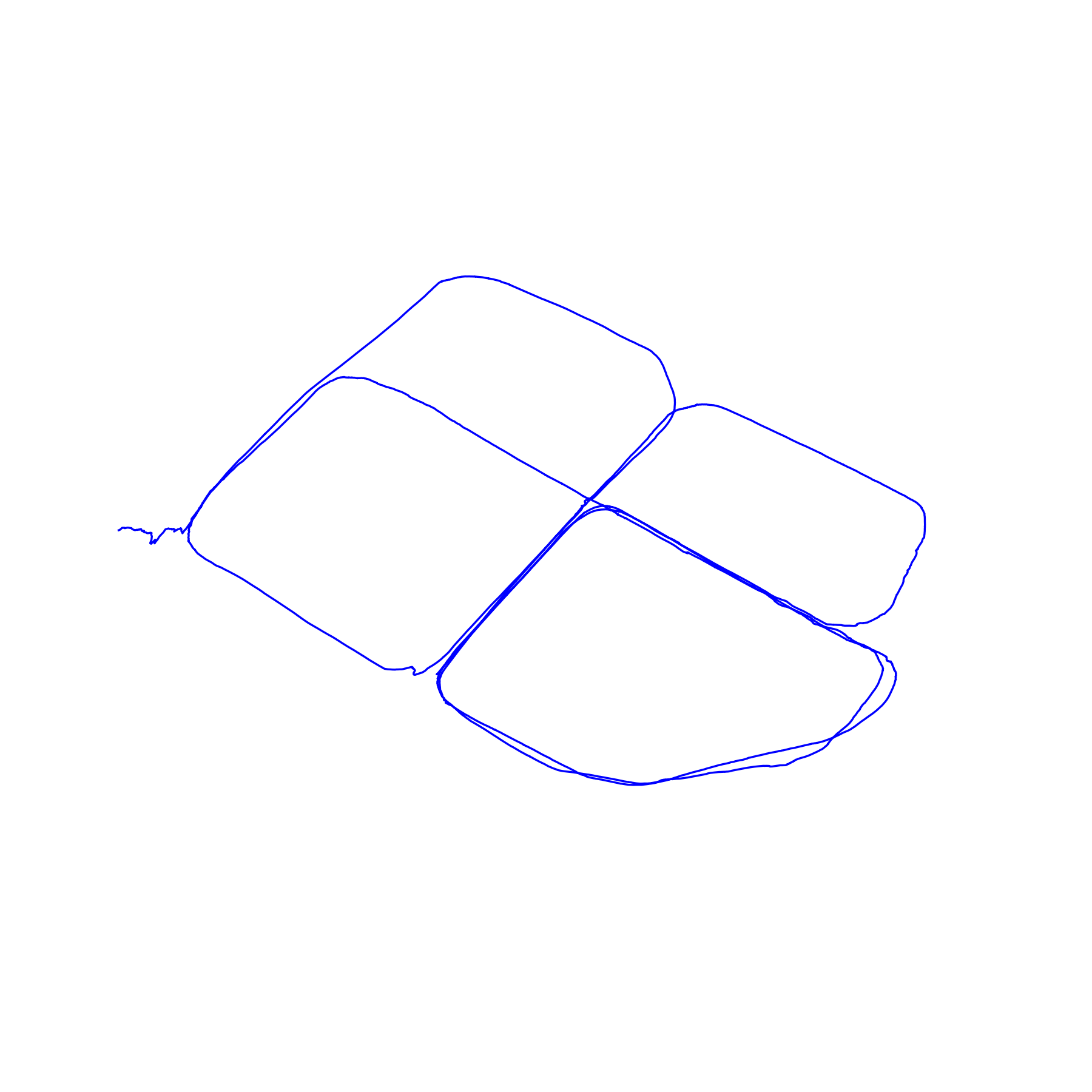}
    \caption{Cubicle}
  \end{subfigure}
  \hfill
  \begin{subfigure}[b]{0.3\textwidth}
    \includegraphics[width=\textwidth]{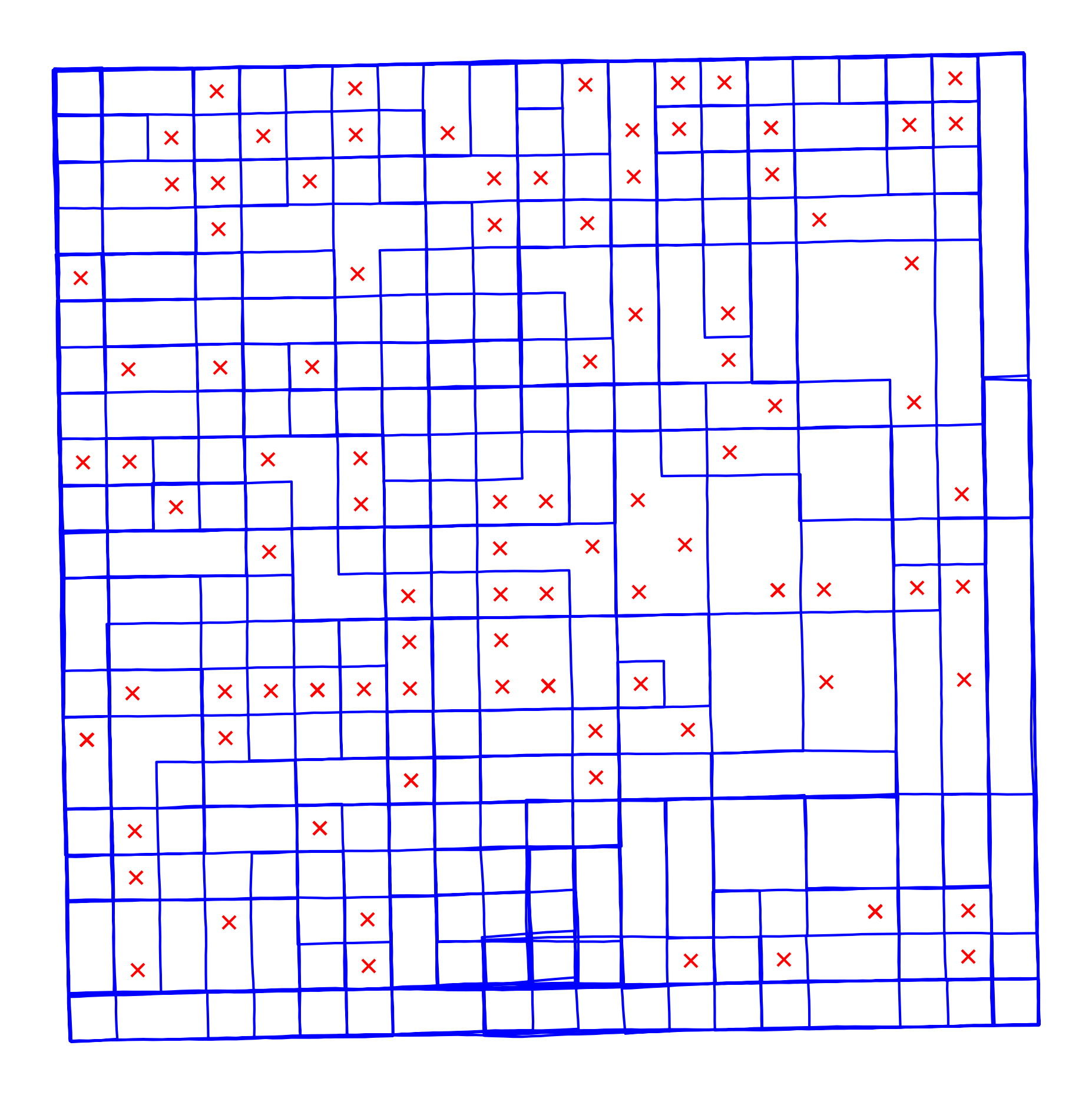}
    \caption{Trees}
  \end{subfigure}
  \hfill
  \begin{subfigure}[b]{0.3\textwidth}
    \includegraphics[width=\textwidth]{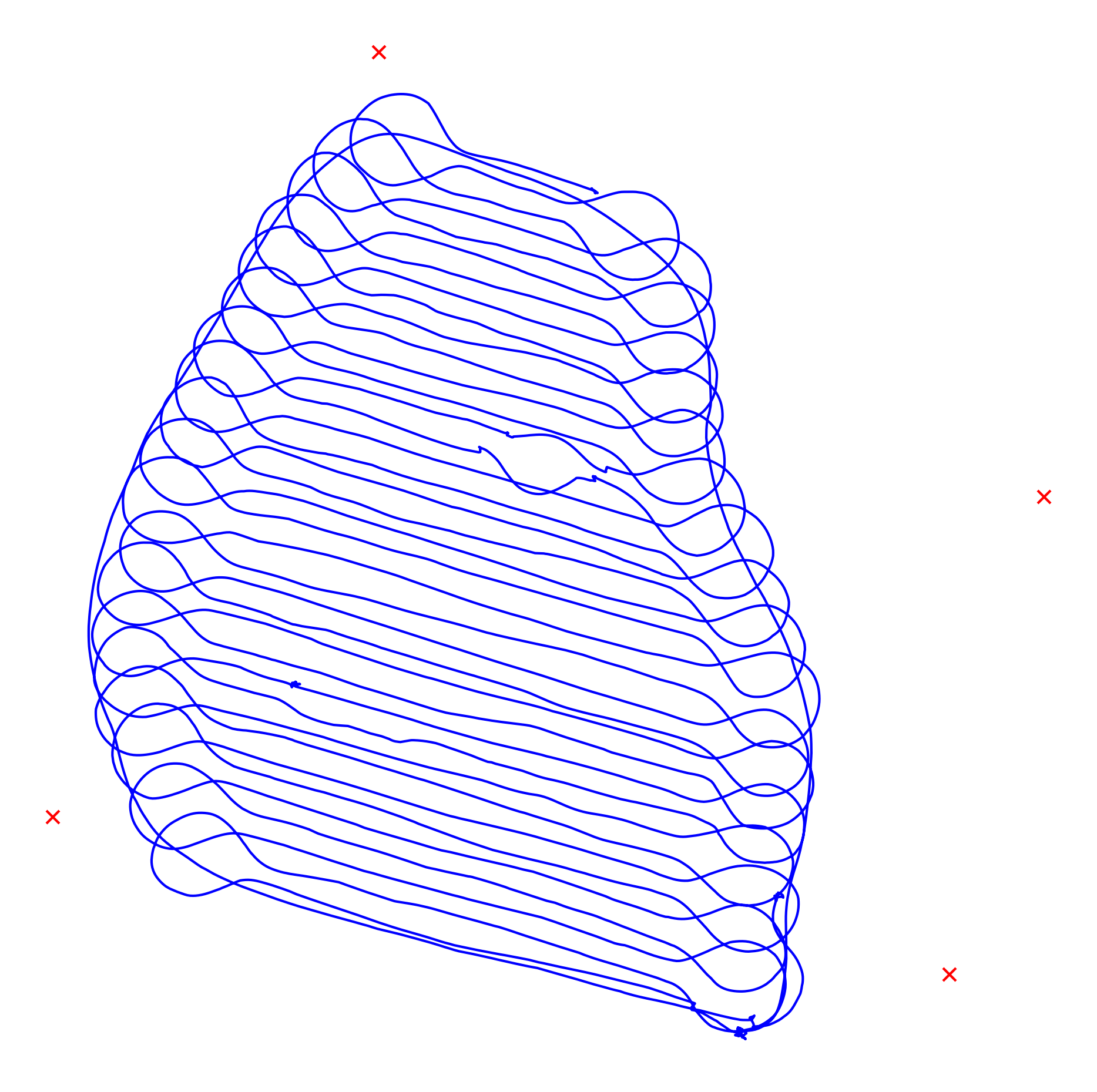}
    \caption{Plaza 1}
  \end{subfigure}

  \vspace{0.5cm}

  \begin{subfigure}[b]{0.3\textwidth}
    \includegraphics[width=\textwidth]{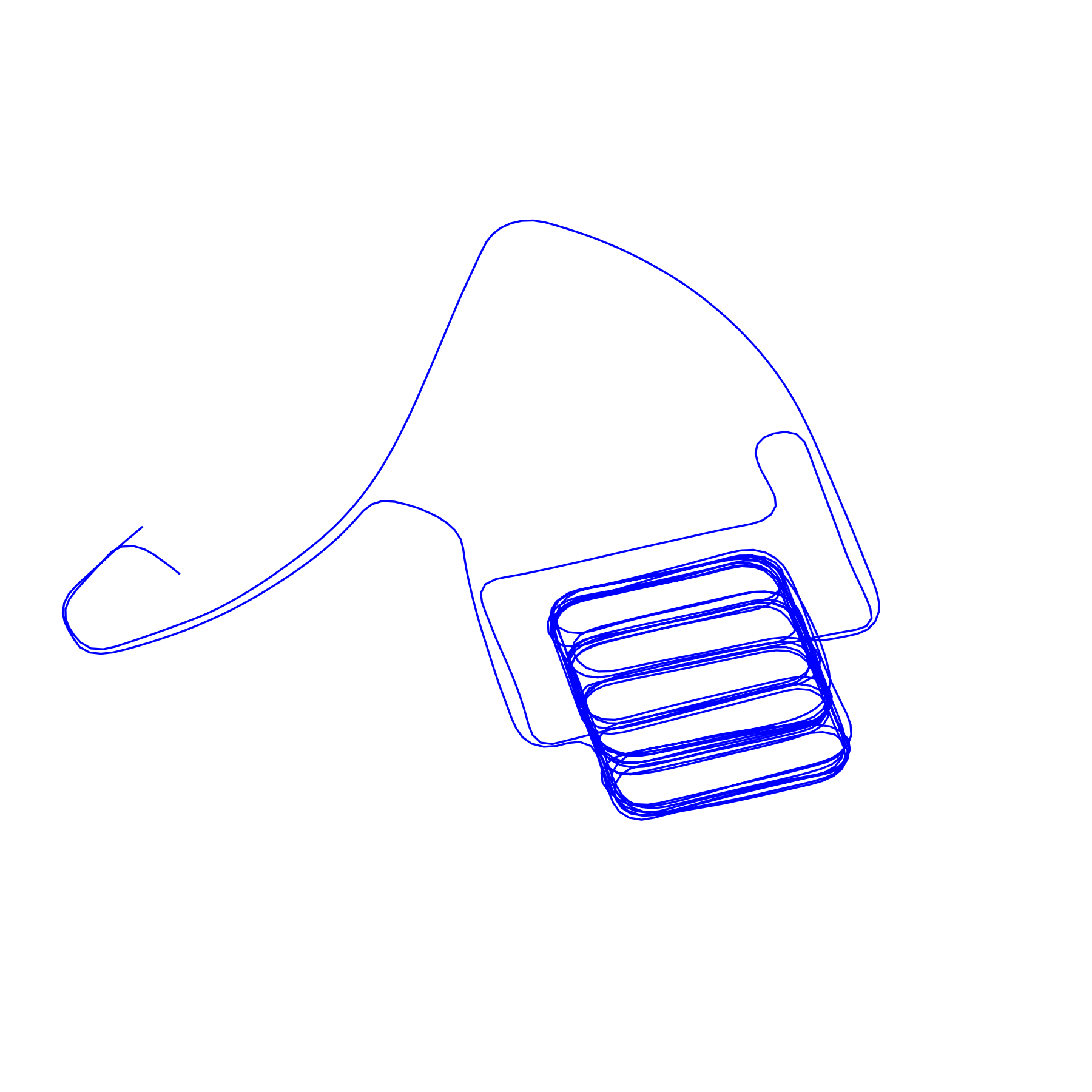}
    \caption{Garage}
  \end{subfigure}
  \hfill
  \begin{subfigure}[b]{0.3\textwidth}
    \includegraphics[width=\textwidth]{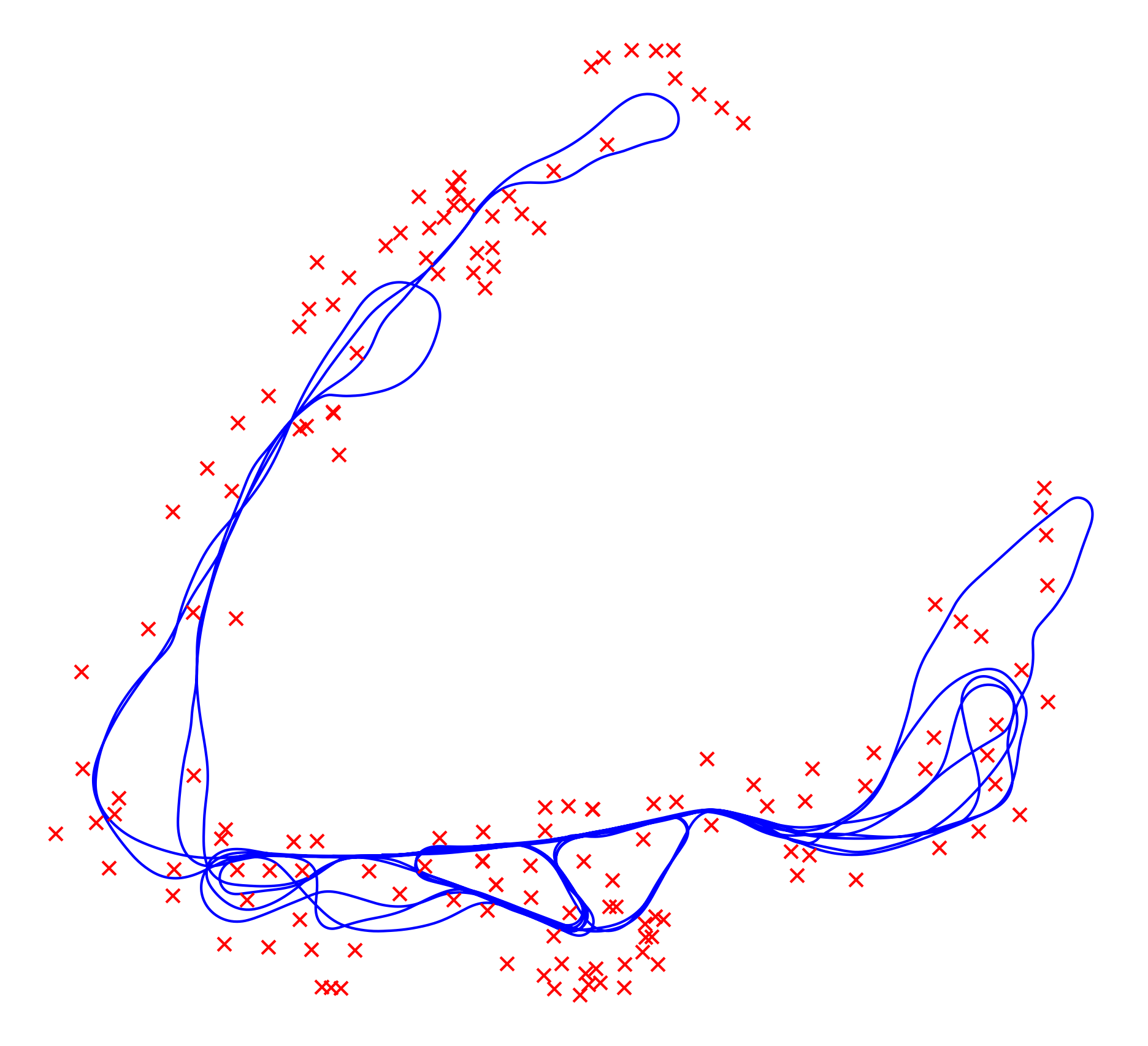}
    \caption{Victoria}
  \end{subfigure}
  \hfill
  \begin{subfigure}[b]{0.3\textwidth}
    \includegraphics[width=\textwidth]{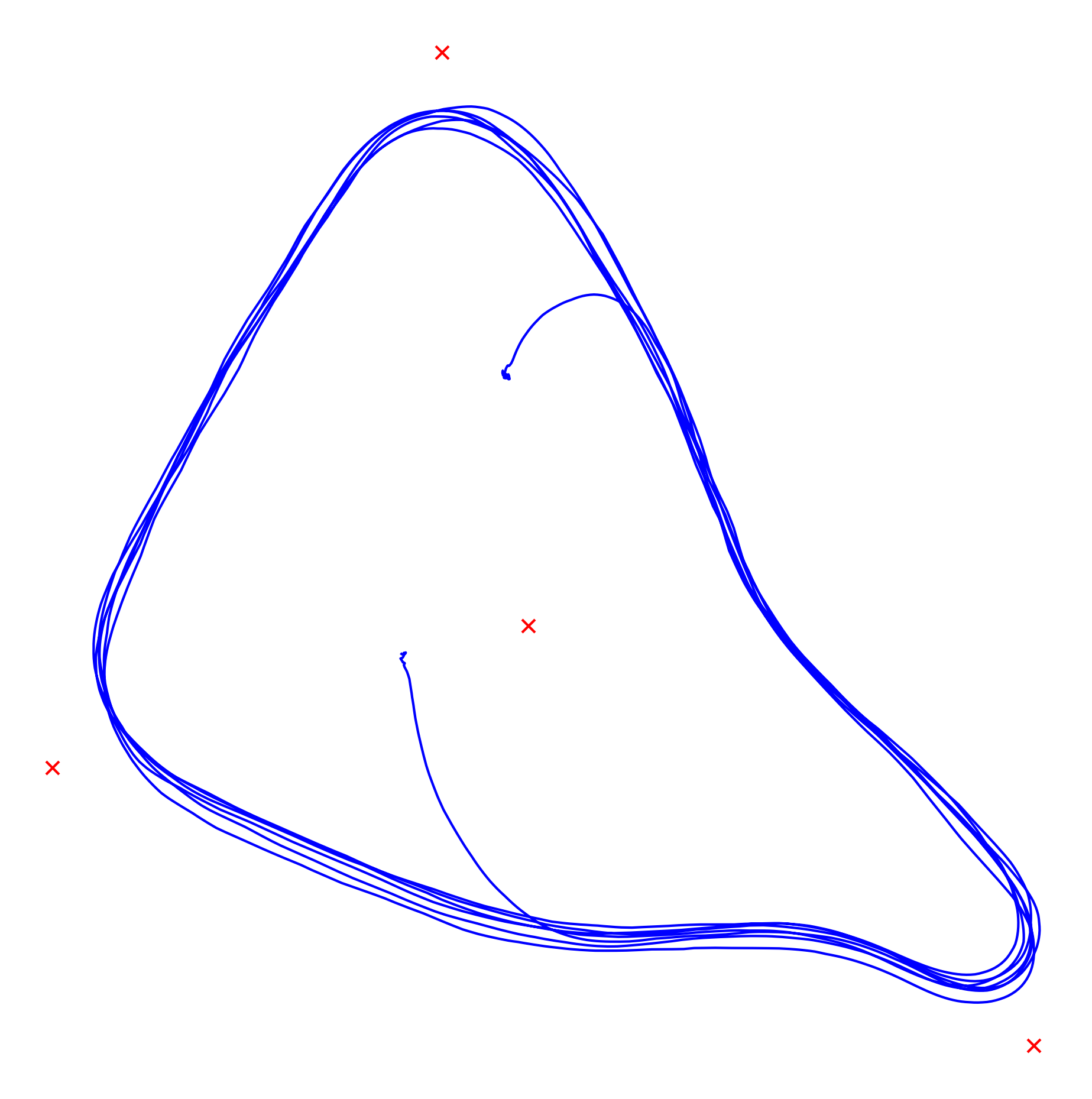}
    \caption{Plaza 2}
  \end{subfigure}

    \caption{Globally optimal solutions recovered by our method on select pose graph optimization, landmark SLAM, and range-aided SLAM benchmarks. Robot trajectories in blue; landmarks (where applicable) in red.}
  
  \vspace{-\baselineskip}
\end{figure*}

\subsection{SLAM problem formulations}
We begin by briefly reviewing the formulations of the three SLAM problem classes considered in our experiments.

\subsubsection{Pose graph optimization}

In {pose graph optimization} (PGO), the goal is to jointly estimate a collection of $n$ unknown poses \( x_1, \dots, x_n \in \mathrm{SE}(d) \) given a set of noisy pairwise relative measurements $\tilde{x}_{ij} = (\tilde{t}_{ij}, \tilde{R}_{ij}) \approx x_i\inv x_j$.  This problem can be represented by a directed graph \( \mathcal{G} = (\mathcal{V}, \mathcal{E}) \) in which the vertices \( \mathcal{V} = [n] \) are in one-to-one correspondence with the unknown poses and the edges \( \mathcal{E} \subseteq \mathcal{V} \times \mathcal{V} \) are in one-to-one correspondence with the available measurements \cite{grisetti2011tutorial}. Assuming the noise models in \eqref{isotropic_Langevin_noise_model} and \eqref{relative_translation_noise_model} for the rotational and translational components of each relative pose measurement (respectively), the corresponding maximum likelihood estimation problem is \cite{rosen2019se}:

\begin{problem}[MLE formulation of PGO]
\label{MLE_formulation_of_PGO_problem}
 \begin{equation}
 \label{equation: non_lifted_PGO_formulation}
\min_{\substack{t_i \in \R^d \\ \R_i \in \SO(d)}}
\sum_{(i,j)\in\mathcal{E}} 
\kappa_{i j} \left \lVert R_j-R_i \tilde{R}_{i j} \right \rVert _F^2 + \tau_{i j} \left \lVert t_j-t_i-R_i \tilde{t}_{i j} \right \rVert_2^2.
\end{equation}
\end{problem}

\subsubsection{Landmark SLAM}

{Landmark SLAM} generalizes pose graph optimization by additionally estimating the positions of $m$ unknown landmarks $l_1, \dotsc, l_m \in \R^d$ in the environment, given an additional set of noisy measurements  $\tilde{l}_{ij} \in \R^d$ of the relative translation from pose $i$ to landmark $j$.  Assuming these relative translation measurements are also sampled from the noise model \eqref{relative_translation_noise_model}, the corresponding maximum likelihood estimation problem can be expressed as \cite{fan2020cpl}:

\begin{problem}[MLE formulation of landmark SLAM]
\label{MLE_formulation_of_landmark_SLAM_problem}
 \begin{equation}
 \label{landmark_SLAM_MLE_formulation_eq}
 \resizebox{\columnwidth}{!}{$\displaystyle
\min_{\substack{t_i\in \R^d \\ 
R_i \in \SO(d)\\
l_j \in \R^d}} 
\!\!
\left \lbrace 
\!\!\!
\begin{split}
&\sum_{(i,j)\in\mathcal{E}} \kappa_{i j} \left \lVert R_j-R_i \tilde{R}_{i j} \right \rVert_F^2  + \tau_{i j} \left \lVert t_j-t_i-R_i \tilde{t}_{i j} \right \rVert_2^2 \\
&\quad + \sum_{(i,j) \in \mathcal{L}} \zeta_{i j} \left \lVert l_j-t_i-R_i \tilde{l}_{i j} \right \rVert_2^2
\end{split}
\right \rbrace
$}
\end{equation}
where $\zeta_{ij} \ge 0$ is the precision of the $ij$-th pose-landmark measurement $\tilde{l}_{ij}$ [cf.\ \eqref{relative_translation_noise_model}], and $\mathcal{L} \subseteq [n] \times [m]$ is the set indexing the available pose-landmark measurements.
\end{problem}

\subsubsection{Range-aided SLAM}

{Range-aided SLAM} (RA-SLAM) likewise generalizes pose graph optimization to the joint estimation of poses and landmarks, but assumes access only to noisy scalar \emph{range} measurements $\tilde{r}_{ij} \in \R$ between pose $i$ and landmark $j$, rather than \emph{relative translation} measurements. Assuming that these range measurements are sampled from the noise model \eqref{range_noise_model}, the corresponding maximum likelihood estimation problem can be expressed as \cite{papalia2024certifiably}:

\begin{problem}[MLE formulation of range-aided SLAM]
\label{MLE_formulation_of_RA_SLAM_problem}
 \begin{equation}
 \label{RA_SLAM_MLE_formulation_eq}
  \resizebox{\columnwidth}{!}{$\displaystyle
\min_{\substack{t_i\in \R^d \\ 
R_i \in \SO(d)\\
l_j \in \R^d}} 
\!\!
\left \lbrace 
\begin{split}
\!\!\!
\!\!\!
&\sum_{(i,j)\in\mathcal{E}} \kappa_{i j} \left \lVert R_j-R_i \tilde{R}_{i j} \right \rVert_F^2  + \tau_{i j} \left \lVert t_j-t_i-R_i \tilde{t}_{i j} \right \rVert_2^2 \\
&\quad + \sum_{(i,j) \in \mathcal{R}} \frac{1}{\sigma_{ij}^2}\left( \lVert l_j - t_i \rVert_2 - \tilde{r}_{ij}\right)^2
\end{split}
\right \rbrace
$}
\end{equation}
where $\sigma_{ij} > 0$ is the standard deviation of the $ij$-th ranging measurement $\tilde{r}_{ij}$ [cf.\ \eqref{range_noise_model}], and $\mathcal{R} \subseteq [n] \times [m]$ is the set indexing the available range measurements.
\end{problem}

\subsection{Lifted certifiable formulations}

In this subsection, we apply the framework of Section \ref{Certifiable_factor_graph_optimization_section} to derive the lifted certifiable formulations corresponding to Problems \ref{MLE_formulation_of_PGO_problem}--\ref{MLE_formulation_of_RA_SLAM_problem}.  Concretely, these lifted problems are obtained by replacing each variable and factor appearing in the original MLE formulations with its corresponding lifted counterpart (cf.\ Section \ref{common_lifted_types_section}).  This substitution produces:

\begin{problem}[Lifted formulation of PGO]
\label{lifted_formulation_of_PGO_problem}
 \begin{equation}
 \label{BM_factored_SDP_for_PGO}
\min_{\substack{t_i \in \R^p \\ Y_i \in \Stiefel(d,p)}}
\sum_{(i,j)\in\mathcal{E}} 
\kappa_{ij} \left \lVert Y_j-Y_i \tilde{R}_{ij} \right \rVert_F^2 + \tau_{ij} \left \lVert t_j- t_i - Y_i \tilde{t}_{ij} \right \rVert_2^2
\end{equation}
\end{problem}

\begin{problem}[Lifted formulation of landmark SLAM]
\label{lifted_formulation_of_landmark_SLAM_problem}
 \begin{equation}
 \label{BM_factored_SDP_for_landmark_SLAM}
  \resizebox{\columnwidth}{!}{$\displaystyle
\min_{\substack{t_i\in \R^p \\ 
Y_i \in \Stiefel(d,p)\\
l_j \in \R^p}} 
\!\!
\left \lbrace 
\!\!
\begin{split}
\!\!\!
\!\!\!
&\sum_{(i,j)\in\mathcal{E}} \kappa_{ij} \left \lVert Y_j - Y_i \tilde{R}_{i j} \right \rVert_F^2  + \tau_{ij} \left \lVert t_j - t_i - Y_i \tilde{t}_{ij} \right \rVert_2^2 \\
&\quad + \sum_{(i,j) \in \mathcal{L}} \zeta_{i j} \left \lVert l_j - t_i - Y_i \tilde{l}_{ij} \right \rVert_2^2
\end{split}
\right \rbrace
$}
\end{equation}
\end{problem}

\begin{problem}[Lifted formulation of range-aided SLAM]
\label{lifted_formulation_of_RA_SLAM_problem}
 \begin{equation}
 \label{BM_factored_SDP_for_RA_SLAM}
  \resizebox{\columnwidth}{!}{$\displaystyle
\min_{\substack{t_i \in \R^p \\ 
Y_i \in \Stiefel(d,p)\\
l_j \in \R^p \\
s_{ij} \in S^{p-1}
}}
\!\!
\left \lbrace 
\begin{split}
\!\!\!
\!\!\!
&\sum_{(i,j)\in\mathcal{E}} \kappa_{ij} \left \lVert Y_j - Y_i \tilde{R}_{i j} \right \rVert_F^2  + \tau_{ij} \left \lVert t_j - t_i - Y_i \tilde{t}_{i j} \right \rVert_2^2 \\
&\quad + \sum_{(i,j) \in \mathcal{R}} \frac{1}{\sigma_{ij}^2}\left \lVert l_j - t_i - \tilde{r}_{ij} s_{ij} \right \rVert_2^2
\end{split}
\right \rbrace
$}
\end{equation}
\end{problem}

Problems \ref{lifted_formulation_of_PGO_problem}--\ref{lifted_formulation_of_RA_SLAM_problem} are the local optimizations that we will instantiate and solve (via local factor graph-based optimization) at each level of the Riemannian Staircase (Algorithm \ref{Riemannian_Staircase_over_factor_graphs_algorithm}).

\subsection{Comparison with specialized certifiable methods}

In our first set of experiments, we evaluate whether our general certifiable factor graph-based estimation approach reproduces the behavior of state-of-the-art \emph{specialized}, \emph{problem-specific} certifiable estimators for the SLAM formulations described in Problems~\ref{MLE_formulation_of_PGO_problem}--\ref{MLE_formulation_of_RA_SLAM_problem}. To that end, we compare our approach against SE-Sync~\cite{rosen2019se} (for PGO), Landmark SE-Sync (a generalization of SE-Sync to the case of landmark SLAM \eqref{landmark_SLAM_MLE_formulation_eq} provided by the CPL-SLAM library \cite{fan2020cpl}), and CORA~\cite{papalia2024certifiably} (for range-aided SLAM).

\subsubsection{Implementation details} 
\label{certifiable_estimator_comparison_implementation_details_subsection}

All experiments were conducted on a single core of a desktop equipped with an Intel i9 CPU and 64~GB RAM, running Ubuntu~24.04.
To ensure that our experiments exercise the \emph{global} optimization capabilities of these certifiable estimators (including solution verification, saddle escape, and ascent to higher levels of the Riemannian Staircase), we set their initial rank parameters to $p_0 = d$, and construct an initial estimate for each problem instance by randomly sampling a point from the feasible sets of Problems \ref{lifted_formulation_of_PGO_problem}--\ref{lifted_formulation_of_RA_SLAM_problem}.  To enable a fair comparison with our certifiable factor graph-based approach, we run each of SE-Sync, Landmark SE-Sync, and CORA in their \emph{translation-explicit} modes, matching the local optimizer settings (i.e., termination tolerances and $\texttt{max\_iterations}$) and numerical tolerance $\eta$ used in our certifiable factor graph-based estimator for each corresponding problem class, while retaining their default values for all remaining parameters.

Our certifiable factor graph-based estimator (Algorithm~\ref{Riemannian_Staircase_over_factor_graphs_algorithm}) is implemented in C++ using the GTSAM library~\cite{gtsam}, and employs the Levenberg-Marquardt (LM) method to perform the local factor graph optimizations required in line~\ref{factor_graph_local_opt_step}.  For PGO and landmark SLAM problems, we set $\texttt{max\_iterations} = 200$ with termination tolerances $\texttt{relative\_error} = \texttt{absolute\_error} = 10^{-5}$; for RA-SLAM instances, we set $\texttt{max\_iterations} = 300$ with tighter tolerances $\texttt{relative\_error} = \texttt{absolute\_error} = 10^{-8}$.\footnote{Because individual range measurements provide only \emph{partial} information about the relative translations between the involved robot poses and landmark positions, Problem~\ref{lifted_formulation_of_RA_SLAM_problem} is often ill-conditioned, manifesting as poorly-conditioned information matrices and broad flat regions in the loss landscape.  In such cases, a single LM iteration may produce only a marginal decrease in the objective value despite the current estimate being far from stationarity. To account for this, we adopt tighter termination tolerances for RA-SLAM instances to enable local optimization to effectively escape these flat regions.}  To test the positive semidefiniteness of the certificate matrix $S$ constructed in line~\ref{parallel_computation_of_adjoint_diagonal_blocks_step} of Algorithm~\ref{Riemannian_Staircase_over_factor_graphs_algorithm}, we set a fixed numerical tolerance $\eta = 10^{-3}$ for PGO and landmark SLAM problems.  For RA-SLAM instances, we instead employ an adaptive tolerance $\eta$ defined as:
\begin{equation}
\label{numerical_PSD_test}
  \eta \triangleq \min\!\bigl\{\eta_{\max},\, \max\{10^{-5}\, f^\star,\, \eta_{\min}\}\bigr\},
\end{equation}
where $\eta_{\min} = 10^{-7}$, $\eta_{\max} = 10^{-1}$, and $f^\star$ denotes the objective value at the stationary point $Y^\star$ returned by the local factor graph optimization; the certificate is declared valid if $S + \eta I \succeq 0$.  This adaptive tolerance strategy follows that of CORA, and we adopt exactly the same parameter settings as in the baseline CORA experiments.  
Finally, since the SDP relaxation associated with the RA-SLAM problem \eqref{RA_SLAM_MLE_formulation_eq} is known to be slightly loose in practice, for RA-SLAM problems we follow the approach of CORA and perform a final \emph{local} optimization on the rounded feasible estimate recovered from the SDP to obtain a final \emph{refined} estimate, together with a certified \emph{lower bound} on the optimal value for Problem \ref{MLE_formulation_of_RA_SLAM_problem}.

\subsubsection{Experimental results and analysis} 
Results for this experiment are shown in Table~\ref{Table: random_initialization_experiments}.

Across all problem instances, our certifiable factor graph optimization approach faithfully reproduces the behavior of current state-of-the-art \textit{problem-specific} certifiable estimators, recovering estimates with objective values matching those obtained by these specialized methods (up to numerical precision). Consistent with prior work \cite{rosen2019se, fan2020cpl, papalia2024certifiably}, our approach recovers \textit{certifiably globally optimal} solutions for all tested PGO and landmark SLAM instances, and \textit{certifiably near-optimal} solutions for the RA-SLAM test cases (after rounding and refinement all solutions are at most 13\% suboptimal, and all but one are within 6\%). This behavior is expected, as both our approach and the specialized certifiable methods solve the \textit{same} underlying SDP relaxations.

Runtime differences between our approach and the specialized certifiable estimators are primarily attributable to differences in the local optimization methods used to solve the underlying Burer-Monteiro factored Shor relaxations~\eqref{BM_factored_SDP_for_PGO}--\eqref{BM_factored_SDP_for_RA_SLAM}. Our implementation employs GTSAM's general-purpose Levenberg-Marquardt solver, which recomputes and factors Jacobians \textit{in each iteration}, while the problem-specific certifiable estimators all employ specialized Riemannian trust-region methods that exploit the quadratic form of the objective and the geometry of the feasible set, together with a preconditioned conjugate gradient method for computing update steps, to avoid the need to \textit{explicitly} relinearize the objective in each iteration (cf.~\cite[Sec.~5.1.3]{rosen2019se}). Accordingly, our implementation is generally slower, due to the use of a general-purpose solver rather than a specialized, structure-exploiting technique. Importantly, this difference is \textit{not} an inherent limitation of our proposed approach: incorporating such structure-exploiting Riemannian optimization techniques within our factor graph-based framework would yield comparable computational performance. Moreover, despite using a less efficient local optimizer, the computational cost of our implementation is already well within the range required for practical use, and indeed is comparable with that of current state-of-the-art factor graph-based \textit{local} optimization methods widely employed in robotics and computer vision, as we demonstrate in the following subsection.

\subsection{Comparison with local factor graph-based optimization methods}

In our second set of experiments, we evaluate whether our certifiable factor graph-based approach preserves the computational efficiency of current state-of-the-art \emph{local} factor graph-based optimization methods. To do so, we compare our approach with GTSAM's Levenberg-Marquardt optimizer applied directly to the original SLAM MLEs (Problems \ref{MLE_formulation_of_PGO_problem}--\ref{MLE_formulation_of_RA_SLAM_problem}).

\subsubsection{Implementation details}

To enable a fair comparison, we employ identical $\texttt{max\_iterations}$ and termination criteria for the LM optimizers applied directly to the SLAM MLEs (Problems \ref{MLE_formulation_of_PGO_problem}--\ref{MLE_formulation_of_RA_SLAM_problem}) and those used within Algorithm~\ref{Riemannian_Staircase_over_factor_graphs_algorithm}, ensuring that any observed differences in performance arise \emph{solely} from algorithmic strategy rather than implementation details.  We again set the initial relaxation rank in Algorithm \ref{Riemannian_Staircase_over_factor_graphs_algorithm} to $p_0 = d$, and initialize both methods using the \emph{odometric initialization} for the pose variables (obtained by concatenating the relative pose measurements between successive poses along the robot's trajectory), and the \emph{ground truth} values for the landmark variables.  Note that the latter provides a substantially \emph{higher-quality} initialization than what is typically available in real-world landmark SLAM applications, and therefore places the local optimization methods in a particularly favorable regime.\footnote{Indeed, for RA-SLAM the fact that range measurements  provide only partial information about landmark positions means that even constructing good \emph{initializations} for Problem \ref{MLE_formulation_of_RA_SLAM_problem} can be highly nontrivial in practice \cite{PapaliaSCORE2023}.}

\subsubsection{Experimental results and analysis}  

Results for this experiment are shown in Table \ref{Table: odom_initialization_experiments}.

These results confirm that our certifiable factor graph approach preserves the computational efficiency of current state-of-the-art \emph{local} factor graph-based optimization methods (up to small constant factors), while additionally enabling the recovery of \emph{certifiably optimal or near-optimal} solutions.  At a high level, our approach can be viewed as an \emph{adaptive extension} of standard {local} factor graph optimization.  To see this, consider the initial level of the Riemannian Staircase (Algorithm \ref{Riemannian_Staircase_over_factor_graphs_algorithm}): for $p = d$, the local optimizations \eqref{BM_factored_SDP_for_PGO}--\eqref{BM_factored_SDP_for_RA_SLAM} performed in line \ref{factor_graph_local_opt_step} are essentially \emph{identical} to the original MLEs \eqref{equation: non_lifted_PGO_formulation}--\eqref{RA_SLAM_MLE_formulation_eq} for PGO and landmark SLAM, and are closely related for RA-SLAM.  Consequently, for cases in which the provided initialization is sufficient for local optimization to recover a certifiable \emph{global} minimizer (indicated by a checkmark in the rightmost column of Table \ref{Table: odom_initialization_experiments}),  Algorithm \ref{Riemannian_Staircase_over_factor_graphs_algorithm} terminates at this initial level after performing a \emph{single} local optimization of the same computational cost as the MLE, followed by a single optimality verification step with only modest additional overhead \cite{rosen2022accelerating}.  In contrast, for instances in which the provided initialization leads to a \emph{suboptimal} solution (\texttt{MIT}, \texttt{Torus}, \texttt{Cubicle}, \texttt{Rim}, \texttt{Victoria}, and all RA-SLAM instances except \texttt{Goats~15}, where suboptimality is evident from GTSAM's objective values substantially exceeding the certified refined values), our method does incur additional cost when ascending to higher levels of the Riemannian Staircase, which entails performing \emph{multiple} local optimization and verification operations in sequence.
However, as expected, the total computational cost of these operations scales approximately \emph{linearly} in the \emph{terminal level} of the Staircase, and thus remains within a \emph{small constant multiple} of the cost of a single \emph{local} optimization applied to the  MLE. Taken together, these observations reveal that our approach behaves identically to standard \emph{local} factor graph optimization when the provided initialization is sufficient, and only invokes the full machinery of the Riemannian Staircase (for \emph{global} optimization) when global optimality cannot be certified at the initial level.  These results also highlight that, while certifiable estimators do not \emph{require} a high-quality initialization to recover globally optimal solutions, they can nevertheless still benefit computationally from informative initializations whenever these are available.

\subsection{Discussion}
\label{ExpDisc}

These results confirm that our certifiable factor graph optimization framework provides a practically effective unification of the certifiable estimation and factor graph-based \emph{local} optimization paradigms. In particular, our approach recovers solutions with \emph{identical} objective values and certified suboptimality bounds as \emph{specialized}, \emph{problem-specific} certifiable estimators, while preserving the high computational efficiency of state-of-the-art factor graph-based \emph{local} optimization methods.

Crucially, this unification is achieved within a general-purpose factor graph framework, substantially reducing the implementation effort required to develop certifiable estimators. At present, designing and deploying such estimators is a laborious and highly specialized undertaking: practitioners must derive problem-specific SDP relaxations, construct corresponding Burer-Monteiro factorizations, develop custom Riemannian optimization algorithms tailored to the geometry of each problem's feasible set, and then integrate these components within a Riemannian Staircase procedure. Each of these steps requires substantial expertise in convex analysis, differential geometry, and numerical optimization, and must typically be repeated \emph{ab initio} for each new problem class.

In contrast, assembling a certifiable estimator using our certifiable factor graph framework is quite straightforward. A practitioner familiar with the factor graph abstraction need only replace the original variable and factor types in a factor graph model of the target estimation task with their lifted certifiable counterparts; the remainder then follows directly from the framework proposed in this paper. In particular, no specialized expertise in convex optimization, semidefinite programming, or differential geometry is required, and no custom solver development is needed.  This reduction in implementation effort transforms certifiable estimation from a \emph{specialized}, \emph{months-long} development effort into a straightforward task that can be completed in \emph{hours} using standard factor graph libraries and workflows.

\section{Conclusions}
\label{conclusion_section}

In this paper, we showed that factor graph optimization and certifiable estimation can be synthesized into a unified theoretical and algorithmic framework for \emph{certifiable factor graph optimization}. The key insight enabling this synthesis is that the core constructions underpinning certifiable estimators, namely Shor relaxation and Burer-Monteiro factorization, inherit a natural factor graph structure from the original estimation task. This insight enables certifiable  estimators to be designed and implemented using the same factor graph-based modeling abstractions and software libraries already ubiquitously employed throughout robotics and computer vision, eliminating the need for custom semidefinite programming formulations and hand-designed, problem-specific Riemannian optimization algorithms. The resulting framework provides a general and systematic methodology for constructing certifiable estimators directly from factor graph models, substantially reducing the implementation effort and domain expertise required to deploy certifiable estimators in practice. Moreover, by integrating certifiable estimation into the factor graph paradigm, this work enables the direct application of over two decades of advances in high-performance factor graph-based inference, particularly for large-scale and real-time estimation \cite{dellaert2017factor}. This connection provides a clear path forward towards improved scalability, computational efficiency, and ease of use, all of which are essential for establishing certifiable estimation as a practical, general-purpose tool for robust robotic state estimation.

\appendices

\balance
\bibliographystyle{IEEEtran}
\bibliography{IEEEabrv, ref/ref}

\end{document}

%% file: define.tex
\usepackage{subcaption}
\usepackage{tabularx} 

\newcommand{\cp}[1]{\ifmmode {\mathcal{#1}}\else ${\mathcal{#1}}$\fi}
\usepackage{array}
\newcolumntype{P}[1]{>{\centering\arraybackslash}p{#1}}

\usepackage{nccmath}

\def\credrev2{\textcolor{red}}
\def\credrev{\textcolor{red}}

\definecolor{darkgreen}{rgb}{0., 0.4, 0.}
\definecolor{amber}{rgb}{1.0, 0.49, 0.0}
\definecolor{orange}{rgb}{1.0, 0.4, 0.0}